\pdftrailerid{}

\documentclass[10pt,twocolumn,letterpaper]{article}

\usepackage[pagenumbers]{cvpr}%

\usepackage{amsmath,amssymb,mathtools}
\usepackage{amsthm}
\newtheorem{proposition}{Proposition}
\usepackage{booktabs}
\usepackage{graphicx}
\usepackage{xcolor}
\usepackage{enumitem}
\usepackage{textcomp}
\usepackage{multirow}
\usepackage{array}

\DeclareMathOperator{\LSE}{LSE}
\newcommand{\Ntr}{N_{\mathrm{tr}}}
\newcommand{\lam}{\lambda}

\definecolor{cvprblue}{rgb}{0.21,0.49,0.74}
\usepackage[pagebackref,breaklinks,colorlinks,allcolors=cvprblue]{hyperref}

\def\confName{CVPR}
\def\confYear{2027}

\title{Spread Too Thin: How Higher Resolution Dilutes\\a Vision Transformer's Attention, and a Training-Free Rescaling}

\author{Neil D.\ B.\ Bruce\\University of Guelph, Guelph, Ontario, Canada\\{\tt\small brucen@uoguelph.ca}}

\hypersetup{pdftitle={Spread Too Thin: How Higher Resolution Dilutes a Vision Transformer's Attention, and a Training-Free Rescaling},pdfauthor={Neil D. B. Bruce},pdfsubject={},pdfkeywords={},pdfcreator={},pdfproducer={}}
\begin{document}
\twocolumn[{%
\renewcommand\twocolumn[1][]{#1}%
\maketitle
\vspace{-1.2em}
\begin{center}
\captionsetup{type=figure}
\includegraphics{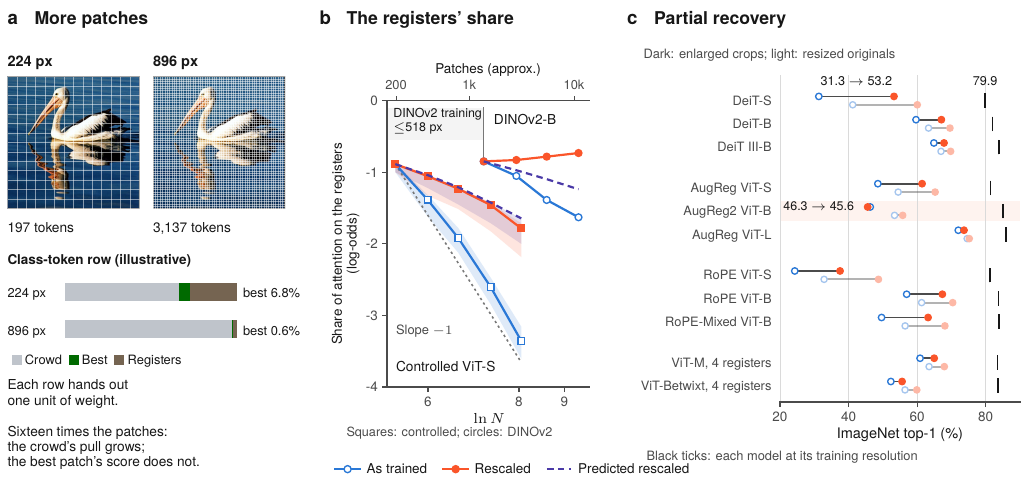}
\captionof{figure}{\textbf{Rescaling every score recovers part of what a vision transformer loses when more patches thin its attention.}
(a) The same image with more patches; token counts include the class token. Unit bars are illustrative, with additional registers.
(b) Measured class-token register shares; the controlled median and band span 3{} seeds, not a confidence interval.
Dashed curves predict rescaled shares from training-resolution scores. The top patch-count scale is approximate;
the horizontal data use $\ln N$. Rescaling multiplies every score by $\ln N/\ln\Ntr$.
(c) All 11{} public checkpoints at sixteen times the patches, on the full validation set.
Pointwise 95\% image-bootstrap intervals are mostly smaller than the marks; paired gain intervals are in \cref{tab:checkpoints}.
No model reaches its own training-resolution tick. AugReg2 ViT-B loses on enlarged crops, despite gaining on resized originals.}
\label{fig:teaser}
\end{center}
\vspace{0.6em}
}]
\begin{abstract}
A vision transformer read at a higher resolution than it was trained at spreads each attention row over many more
patches, yet every row must still hand out exactly one unit of weight. A law of softmax attention predicts that a
growing crowd of bounded scores dilutes the attention on any patch or register whose score stays fixed. The same law
gives a remedy that needs no training: multiply every score, the sink's included, by the ratio of the logarithms of the
test and training token counts. Given sufficient margins at the training resolution, this keeps the attention on a
match when one is present, and on a sink such as a register when none is, though trained models need not meet those
margins. We test these predictions in public vision
transformers and in controlled models read far above their training token count. Attention spreads as predicted. Registers lose their
share, though production models compensate in part, and the rescaling holds the share of DINOv2's registers nearly
level. Applied at test time without training, the rescaling recovers part of the lost accuracy in nearly every model,
with absolute or rotary positions and with registers, and on images with native detail. The calibrated multiplier
rises with the logarithmic ratio on average and varies across models. Its rise is steeper than the square-root rule
used in image generators; in controlled models the log ratio is also more accurate, while in the public models the
two rules' accuracy difference remains unresolved. The rescaling does not restore the attention a model learned, and why it
helps, beyond countering the dilution, remains open.
\end{abstract}
\section{Introduction}\label{sec:intro}

A vision transformer~\citep{dosovitskiy2021vit} reads an image as a grid of patches, and in every attention row a query
hands out one unit of weight over all of them. Read a model trained at 224~px on an 896~px image and each row spreads
over sixteen times as many patches. Accuracy falls steeply: DeiT-S~\citep{touvron2021training} drops from
79.9\% top-1 at its training size to 41.2\% on ImageNet validation images resized to 896~px. Practice has
two separate answers. Image generators and resolution-flexible models rescale attention scores when the token count
changes~\citep{jin2023training,gao2024lumina,wang2024fitv2,leroy2024winwin}, and registers provide extra tokens for
internal computation and reduce artifacts in patch features~\citep{darcet2024registers,jiang2025registers}.

A recent law of softmax attention connects the two~\citep{bruce2026sinks}. A head has two jobs: to put its weight on the
key that matters when there is one, and on a sink when there is none. While content scores stay in a range that does
not grow with the number of tokens, one sink shared by both kinds of query cannot keep both jobs as the number of keys
$N$ grows, and multiplying every score, the sink's included, by $\lam=\ln N/\ln\Ntr$ keeps both under separated margins.
For higher-resolution vision transformers, we test the law's bounded-score growth predictions
(\cref{sec:predictions}): the crowd's log-sum grows like $\ln N$, and a fixed-score sink's log-odds fall like $-\ln N$.
A gate outside the softmax keeps its share when its input is fixed; on a crowd that repeats its training scores, the
rescaled share follows from the training row.

We test these predictions where practitioners meet them: in eleven public ImageNet-1k vision transformers from three
families, with absolute or rotary positions and with registers; in DINOv2 with registers~\citep{oquab2024dinov2}; and in
controlled models, all read at up to sixteen times their training token count, on enlarged crops, on resized originals
and on images with native detail.

Attention spreads as predicted, though in pretrained models the crowd grows somewhat more slowly than the patch count.
Zero-value registers lose their share at the predicted rate, production registers lose it more slowly, and the
rescaling holds DINOv2's register share nearly level. Applied at test time without training, the rescaling recovers part
of the lost accuracy in 10{} of 11{} checkpoints on enlarged crops and in all 11{} on resized
originals; at the largest evaluated resolution no public classifier returns to its training-resolution accuracy, and
one model loses accuracy on enlarged crops. Across the calibrated public models, the selected multiplier rises with
$\lam$ at a slope of 0.945{} (95\% interval [0.635, 1.266]), which excludes the square-root rule's
0.454, though it varies from model to model. The public models' held-out accuracy difference between the
log-ratio and square-root rules remains unresolved. In the calibrated public models, larger median
margins at the training size are associated with smaller selected multipliers, consistent with the direction suggested
by the head-level analysis (\cref{sec:multiplier-head}).

\paragraph{Contributions.}
\begin{itemize}[leftmargin=*,itemsep=1pt,topsep=2pt]
\item \emph{A test of a softmax law's predictions in public vision transformers} (measured):
eleven ImageNet-1k checkpoints and controlled models, on two input routes and a native-detail subset.
\item \emph{Register dilution against resolution in production models} (measured): the predicted rate in controlled
models, partial compensation in production ones, and the share under rescaling predicted from training-resolution scores.
\item \emph{A held-out comparison of the multiplier rules in use} (measured): the log ratio, its square root, FiTv2's and
YaRN's forms, and a calibrated constant.
\item \emph{Which multiplier preserves the top key's attention weight} (derived, in an idealized setting): the log
ratio for a head that puts half its weight on its top key, less for more decisive heads.
\end{itemize}

\paragraph{What is not new.} Rescaling attention scores with the token count is in use in language models and image
generators (\cref{sec:related}). The law, controlled ViT-S tests and initial pretrained-classifier sweeps are prior
work~\citep{bruce2026sinks}; the sweeps used input normalization different from the checkpoints' prescribed
normalization.
We know of no held-out comparison of these rules on pretrained classifiers; Win-Win tunes a temperature for dense
prediction at high resolution~\citep{leroy2024winwin}. We claim no new architecture or
state-of-the-art result.
\section{Related work}\label{sec:related}

\paragraph{Sinks and registers.} Large vision transformers can repurpose low-information patches to store global image
information; registers provide extra tokens for this computation and reduce artifacts in patch
features~\citep{darcet2024registers}. Registers can also be added at test time by redirecting outlier
activations~\citep{jiang2025registers}. In language models the first tokens act as attention sinks~\citep{xiao2024streamingllm,
sun2024massive}, and designs give attention an explicit place to rest: an extra logit of zero~\citep{miller2023offbyone},
learned sink logits~\citep{openai2025gptoss} and output gates~\citep{qiu2025gated}; concurrent work separates abstention
from noise filtering in value-gated heads~\citep{wang2026abstention}. We measure how the registers' share of attention
changes with resolution, and what rescaling does to it.

\paragraph{Resolution transfer.} Because training crops enlarge objects, a model can gain from testing somewhat above
its training resolution~\citep{touvron2019fixres}, and
models can be trained for many patch sizes~\citep{beyer2023flexivit} or native resolutions~\citep{dehghani2023navit},
with interpolated absolute or rotary positions~\citep{dosovitskiy2021vit,heo2024rotary}. LookHere uses directed
attention masks to improve resolution extrapolation in ViT classifiers and introduces ImageNet-HR for evaluating
high-resolution classifiers~\citep{fuller2024lookhere}. These approaches change training, inputs or positional
attention; we study a scalar multiplier applied to attention scores at test time, which can be combined with them.

\paragraph{Scaling attention with the token count.} Language models multiply attention logits by log-length
factors~\citep{chiang2022overcoming,su2021entropy,bai2023qwen,peng2024yarn,nakanishi2025ssmax}. Image generators
rescale attention when the resolution changes: by $\sqrt{\ln N/\ln\Ntr}$, chosen to hold attention entropy
fixed~\citep{jin2023training,gao2024lumina}, or by $\max\bigl(1,\sqrt{\ln(N/\Ntr)}\bigr)$~\citep{wang2024fitv2}; Win-Win
tunes attention temperatures for dense prediction, motivated by preserving self-attention probabilities at high
resolution~\citep{leroy2024winwin}. ExtraVAR estimates a separate scale for each head to match normalized entropy
across resolutions in visual autoregressive generation~\citep{yan2026extravar}. We derive an odds-preserving multiplier
for an idealized crowd (\cref{sec:multiplier-head}) and compare scalar token-count rules and a calibrated constant on
held-out images of pretrained classifiers.

\paragraph{The law.} That one key needs a lead growing like $\log N$ over $N$ competitors is known~\citep{nakanishi2025ssmax,
velickovic2025softmax}; the law adds the sink's side and the conditions under which the log ratio keeps both
jobs~\citep{bruce2026sinks}. We test its predictions; we do not re-prove it.
\section{What a softmax law predicts at higher resolution}\label{sec:predictions}

\paragraph{The setting.} Take the class token's row in one head of a vision transformer read with $N$ keys: the
patches, the class token and any registers. A key $j$ has score (logit) $s_j$ and weight $p_j=e^{s_j}/\sum_k e^{s_k}$.
Call the keys that do not matter to this row the \emph{crowd}, with combined pull $\LSE=\ln\sum_{j\in\text{crowd}}
e^{s_j}$. At the training resolution $N=\Ntr$; a ViT/16 at 224~px has $\Ntr=197$. The rescaling studied here
multiplies every score before the softmax, the class token's and the registers' included, by
\begin{equation}
\lam=\frac{\ln N}{\ln \Ntr},
\label{eq:rescale}
\end{equation}
which is exactly the identity at the training size and costs nothing beyond the higher-resolution pass itself.

\paragraph{What the law predicts.} We use the sink and rescaling results of Bruce~\citep{bruce2026sinks} and the dilution
consequences given in the supplement, stated here without proof. Every mathematical statement in this paper,
the restated ones included, is also proved in Lean~4; the proofs accompany the paper as ancillary files.

\noindent\fbox{\parbox{\dimexpr\linewidth-2\fboxsep-2\fboxrule\relax}{%
\begin{enumerate}[label=(L\arabic*),leftmargin=*,itemsep=1pt,topsep=2pt]
\item \emph{Dilution}, from the concentration argument of \citep[Prop.~7]{bruce2026sinks}. While the scores stay in a range that does not grow with $N$, the crowd's pull grows like $N$:
$\LSE$ and the attention entropy grow like $\ln N$, while the top score stays bounded.
\item \emph{Sinks are gates}~\citep[Prop.~2]{bruce2026sinks}. A register whose value is zero, a learned sink logit and the extra one of off-by-one
softmax each act as a gate in front of an ordinary softmax. For a fixed sink score and content scores in a range that
does not grow with $N$, the sink's log-odds fall like $-\ln N$; a gate outside the softmax, such as a headwise output
gate, keeps its share as long as its input does not change.
\end{enumerate}}}

\noindent\fbox{\parbox{\dimexpr\linewidth-2\fboxsep-2\fboxrule\relax}{%
\begin{enumerate}[label=(L\arabic*),leftmargin=*,itemsep=1pt,topsep=2pt,start=3]
\item \emph{One sink cannot do two jobs}~\citep[Thms.~4, 8]{bruce2026sinks}. No sink shared by a query with a match and one without keeps both jobs past
a sharp token count while the score range stays fixed. Multiplying every score, the sink's included, by $\lam$ makes both
weights tend to one when the sink's score $b$ (for a group of registers, the largest) exceeds the crowd's bound $s_0$ by more than $\ln\Ntr$
and the match outweighs the sink; a sink left outside the multiplier loses the
guarantee.
\item \emph{What the rescaling keeps}~\citep[Thm.~9]{bruce2026sinks}. Under $\lam$ the sink's odds against the crowd are determined exactly by a power
mean of the crowd's exponentiated scores, so on a crowd that repeats the training scores the share left after rescaling can be
predicted from the scores at the training resolution. For a group of registers we treat them as one sink whose score is the log-sum-exp of theirs, an
approximation that is exact for a single register.
\end{enumerate}}}

\subsection{Which multiplier preserves the top key's attention weight}\label{sec:multiplier-head}
Here preserving a head's decision means preserving the attention weight of its top key, equivalently its odds against
all other keys. The guarantee in (L3) needs margins that trained heads need not meet, so which multiplier suits a trained model is an
empirical question. One idealization says what to expect. Let the head's top key lead $n_{\mathrm{tr}}$ other keys of equal score by $g$ at
the training size, so that its odds against them are $e^{g}/n_{\mathrm{tr}}$. Read with $n$ other keys and every score
multiplied by $c$, the odds become $e^{cg}/n$, and they are unchanged exactly when
\begin{equation}
c=1+\frac{\ln(n/n_{\mathrm{tr}})}{g}.
\label{eq:keep}
\end{equation}
This is the log ratio $\ln n/\ln n_{\mathrm{tr}}$ when $g=\ln n_{\mathrm{tr}}$, that is, when the top key holds half the
weight at the training size. The rescaling as applied counts all $N=n+1$ keys, which changes $\lam$ by less than
0.0015{} (0.10\%) at the token counts used here, so we write $\lam$ for both. A more decisive
head is kept by a smaller multiplier: to first order, the square-root rule $\sqrt{\lam}$ keeps a head with
$g=2\ln n_{\mathrm{tr}}$, which puts all but a fraction $1/(n_{\mathrm{tr}}+1)$ of its weight on its top key. A less decisive head needs more than $\lam$,
and every head is kept by $c=1$ at its training size. A model's best single multiplier should therefore track $\lam$ when
its heads are balanced at their training size, and fall below it when they are decisive. For a crowd that repeats its
training scores, with the same top score strictly above them, the multiplier that preserves the top key's attention
weight lies between $1+\ln(n/n_{\mathrm{tr}})/g_1$ and $1+\ln(n/n_{\mathrm{tr}})/g_\infty$, where $g_1$ and $g_\infty$
are the top key's leads over the crowd's training attention-weighted mean and over its largest score (proved in the
supplement). \Cref{sec:multiplier} tests the trend, the margins at the training size and the
optimum there.

\begin{table}[t]
\centering
\footnotesize
\setlength{\tabcolsep}{2.5pt}
\begin{tabular}{@{}lc>{\centering\arraybackslash}p{2.05cm}c@{}}
\toprule
Quantity & Ref. & Controlled ViT-S & Public\\
\midrule
Crowd's pull, $\LSE$ & 1 & 1.01--1.04$^{\dagger}$ & 0.77--0.91\\
Entropy of the row & 1 & 0.96--0.98$^{\dagger}$ & 0.98--1.00\\
Top score & 0 & 0.25--0.30$^{\dagger}$ & \textminus{}0.05{} to 0.15\\
Register share, log-odds & $-1$ & \textminus{}0.956, \textminus{}0.965, \textminus{}0.958 & \textminus{}0.35{} to \textminus{}0.78\\
Gate share, log-odds & 0 & +0.010, +0.006, +0.016 & --\\
Rescaled register error & (L4) & within 0.09 & within 0.28\\
\midrule
Calibrated multiplier & 1 & $\lam$ beats $\sqrt{\lam}$ in 60/60 & 0.945{} [0.635, 1.266]\\
\bottomrule
\end{tabular}
\caption{\textbf{Growth references and measurements.} Public models: the six primary models for the first three rows,
with slopes per unit log patch count (224 to 896~px); DINOv2 with registers and the two supervised register models for
the register rows, with slopes per unit log total-key count over each model's own ladder. References for the dilution
and sink rows are logarithmic growth coefficients, measured entries finite fitted slopes; the gate reference holds for a
fixed input. The rescaled-register error is the slope error, $|$measured $-$ predicted$|$ per unit of $\ln N$. The multiplier reference is the
balanced flat-head case of \cref{eq:keep}; measured is the slope of the accuracy-optimal $c^*-1$ on $\lam-1$
(0.454{} for the square-root rule). $^{\dagger}$Cited~\citep{bruce2026sinks}.}
\label{tab:predictions}
\end{table}
\section{Models, inputs and protocol}\label{sec:setup}

\paragraph{Models.} We use eleven public ImageNet-1k classifiers from the timm library~\citep{wightman2019timm}, in
three groups. The six primary models use absolute positions and a class token: DeiT-S and
DeiT-B~\citep{touvron2021training}, DeiT~III-B~\citep{touvron2022deit3}, and AugReg ViT-S, AugReg2 ViT-B and AugReg
ViT-L~\citep{steiner2022augreg}. Three models use rotary positions~\citep{heo2024rotary}. Two models have four
registers and average pooling and were trained at 256~px. For
attention measurements without labels we add DINOv2 with registers at three sizes~\citep{oquab2024dinov2}, trained up to
518~px. For controlled comparisons we use the ViT-S/16 models of Bruce~\citep{bruce2026sinks}, trained on the
ImageNet-100 subset~\citep{tian2020cmc} under five attention rules: softmax, zero-value registers, a learned sink, a
headwise output gate, and the gate with a sink, three seeds each.

\paragraph{Inputs.} Each model is read at its training resolution and at about 2, 4, 8 and 16 times its training token
count (224 to 896~px; 256 to 1{,}024~px for the register models), with its own preprocessing and position interpolation.
\emph{Enlarged crops} upsample the training-size crop, adding tokens without adding image detail; \emph{resized
originals} carry real detail; and a \emph{native-detail} subset keeps only images whose originals have full
detail at the evaluated size~\citep{deng2009imagenet,russakovsky2015ilsvrc}. In the public models $N$ counts every key; the supplement gives the controlled models' count. Attention runs in
float32 with the rescaling applied to every score; at the training size the rescaled model is bit-identical to the
original.

\paragraph{Measured keys.} For the class-token diagnostics in the primary models, the log-sum and the top score use
every patch key, including the highest-scoring patch, and exclude the class key; these models have no registers. The
entropy is that of the patch attention after renormalizing its mass over the patch keys. No patch is selected by a
relevance label: ``crowd'' is the theoretical interpretation.

\paragraph{Which rows count.} For the register, sink and gate shares, a pair of a head and a query row counts for a
model if its median share on the option, over 1{,}000 bank images at the training resolution, is at least 0.10. The
selection uses the training resolution only. Models with a class token use its rows; the register models
with average pooling have no class token, so their rows are the patches' rows. Heads that never use the option leave a
model's share undecided, and those cells are kept and reported, not dropped.

\paragraph{Statistics.} On the public resolution ladders, accuracy is measured on all 50{,}000 validation images, with paired bootstrap intervals over
images; pooled intervals resample models, then images, and summarize this panel rather than all vision transformers.
The model is the unit of replication. The supplement lists checkpoint tags and preprocessing.
\section{Attention spreads, and registers lose their share}\label{sec:dilution}

\begin{figure*}[t]
\centering
\includegraphics{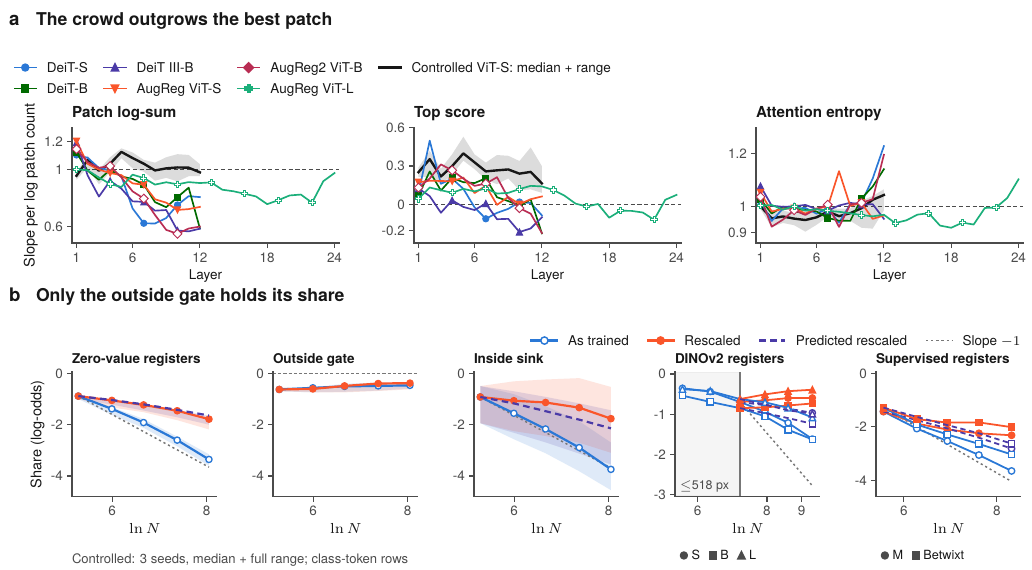}
\caption{\textbf{Both jobs thin as the patches multiply.}
(a) Class-token slopes per unit log patch count for all 6{} primary public checkpoints. The log-sum and
top score use every patch key and exclude the class key; the entropy is that of the patch attention renormalized over
the patches. Reference slopes are 1 for the log-sum and entropy, and 0 for the top score. Black and gray show the median and full range of
9{} controlled ViT-S checkpoints~\citep{bruce2026sinks}.
(b) Share $a$ on the registers, inside sink or outside gate, as log-odds $\ln\bigl(a/(1-a)\bigr)$; for the outside
gate, $a=1-g$ is the suppressed fraction of the head's output. Open marks are as trained;
filled marks rescaled; dashed curves predict the rescaled share from training-resolution scores. Gray dotted
guides have slope $-1$. Controlled curves show the median and full range over 3{} seeds;
these ranges are not confidence intervals. Public curves aggregate medians over images, selected heads, query rows and layers.
DINOv2's shaded range ends at its largest training resolution, 518~px. Shape identifies model size.
The supervised register models use patch rows; all other panels use class-token rows.
Controlled share panels share a vertical scale; the public share scales differ.}
\label{fig:dilution}
\end{figure*}

\paragraph{The crowd outgrows the best patch.} In all 6{} primary models, the
class token's patch log-sum rises at 0.77{} to 0.91{} per unit log patch count, its entropy at 0.98{} to
1.00, while its top score barely moves (\textminus{}0.05{} to 0.15; \cref{fig:dilution}a, \cref{tab:predictions}). The
entropy rises at the predicted rate. The crowd's pull grows somewhat more slowly than the patch count, unlike in the
controlled ViT-S, where it grows at 1.01{} to 1.04~\citep{bruce2026sinks}: pretrained models
compensate in part.
Layer by layer, the crowd's slope is near 1 in the first third of the layers (0.94{} to 1.06{}
by model) and lower in the last third (0.57{} to 0.83), while the entropy's stays near 1 at every
depth (0.92{} to 1.23). These slopes measure spread, not harm: copying every key exactly $k$ times
raises the log-sum and the entropy by exactly $\ln k$ and leaves the top score unchanged, yet the row still gives each
region the same total weight.

\paragraph{Registers lose their share.} In the controlled models, zero-value registers lose their share of the class
token's attention with log-odds slopes of \textminus{}0.956, \textminus{}0.965, \textminus{}0.958{} per unit of $\ln N$, the rate (L2) predicts. A gate outside
the softmax holds its share (+0.010, +0.006, +0.016), while the sink inside the same heads' softmax dilutes (\textminus{}0.933, \textminus{}0.983, \textminus{}0.736):
holding a gate level does not hold what the softmax inside it reads. Production registers lose their share more slowly,
at \textminus{}0.35{} to \textminus{}0.78: they compensate
in part. We write each row's clamped register log-odds as the register group's log-sum-exp minus
a reconstructed crowd term. The fitted slopes satisfy this identity before aggregation; their separately reported
medians need not add. In the 4{} models whose split is stable across the head thresholds, both contribute:
the reconstructed crowd term grows more slowly than the patch count (0.53{} to 0.84{} per unit of
$\ln N$, against 1), and the registers' log-sum-exp rises a little (+0.07{} to +0.09). The
reconstructed crowd term's shortfall is larger than the register-score drift in every model whose split is stable. For the fifth model the split depends on the head threshold, and the supplement gives its slopes only. Inside DINOv2's
training range the share barely falls: from 224 to 336~px its log-odds slopes are \textminus{}0.08{} to \textminus{}0.06,
against \textminus{}0.50{} to \textminus{}0.35{} above 518~px.

\paragraph{Training-score predictions track the rescaled register shares.} Rescaled, DINOv2's register share is held
nearly level (slopes +0.03{} to +0.06). The controlled registers' fall slows to \textminus{}0.426, \textminus{}0.357, \textminus{}0.280. From the
scores at the training resolution alone, (L4) predicts \textminus{}0.363, \textminus{}0.269, \textminus{}0.287{}, agreeing with the measured slopes to within
0.09. In all five production models the prediction is within 0.28. The learned sink has no selected heads in most controlled seeds, so its shares are left undecided.
\section{The rescaling recovers accuracy without training}\label{sec:accuracy}

\begin{table*}[t]
\centering
\footnotesize
\setlength{\tabcolsep}{3.4pt}
\begin{tabular}{@{}lrcccccc@{}}
\toprule
& & Training-size & \multicolumn{3}{c}{16$\times$ tokens: top-1 and gain} & \multicolumn{2}{c}{Share of loss recovered}\\
\cmidrule(lr){4-6}\cmidrule(l){7-8}
Model & Params & top-1 (\%) & Crops: trained $\to$ rescaled & Crops: gain [95\% CI] & Originals: gain [95\% CI] & Crops & Originals\\
\midrule
\multicolumn{8}{@{}l}{\emph{Absolute positions, class token}}\\
DeiT-S & 22M & 79.9 & 31.3 $\to$ 53.2 & +21.9 [21.5, 22.3] & +18.8 [18.4, 19.2] & 0.45 & 0.49\\
DeiT-B & 87M & 82.0 & 59.7 $\to$ 67.1 & +7.4 [7.1, 7.7] & +6.2 [6.0, 6.5] & 0.33 & 0.33\\
DeiT III-B & 87M & 83.8 & 64.9 $\to$ 67.8 & +3.0 [2.7, 3.2] & +2.8 [2.5, 3.1] & 0.16 & 0.17\\
AugReg ViT-S & 22M & 81.4 & 48.5 $\to$ 61.4 & +12.9 [12.5, 13.3] & +10.8 [10.4, 11.2] & 0.39 & 0.40\\
AugReg2 ViT-B & 87M & 85.1 & 46.3 $\to$ 45.6 & \textminus{}0.8 [\textminus{}1.2, \textminus{}0.4] & +2.4 [2.0, 2.8] & \textminus{}0.02 & 0.08\\
AugReg ViT-L & 304M & 85.8 & 72.0 $\to$ 73.7 & +1.7 [1.4, 1.9] & +0.6 [0.3, 0.9] & 0.12 & 0.05\\
\multicolumn{8}{@{}l}{\emph{Rotary positions}}\\
RoPE ViT-S & 22M & 81.2 & 24.3 $\to$ 37.5 & +13.2 [12.8, 13.6] & +15.9 [15.5, 16.3] & 0.23 & 0.33\\
RoPE ViT-B & 86M & 83.7 & 56.9 $\to$ 67.4 & +10.4 [10.1, 10.8] & +9.1 [8.7, 9.4] & 0.39 & 0.40\\
RoPE-Mixed ViT-B & 86M & 83.8 & 49.6 $\to$ 63.2 & +13.6 [13.2, 14.0] & +11.6 [11.2, 12.0] & 0.40 & 0.42\\
\multicolumn{8}{@{}l}{\emph{Registers, average pooling}}\\
ViT-M, 4 registers & 39M & 83.5 & 60.9 $\to$ 65.0 & +4.1 [3.8, 4.4] & +4.4 [4.2, 4.7] & 0.18 & 0.22\\
ViT-Betwixt, 4 registers & 60M & 83.7 & 52.3 $\to$ 55.6 & +3.3 [2.9, 3.6] & +3.4 [3.1, 3.8] & 0.10 & 0.13\\
\midrule
\multicolumn{4}{@{}l}{\emph{Pooled, six primary models}} & +7.68{} [2.10, 14.31] & +6.94{} [2.51, 12.44] & 0.23$^{\ast}$ & 0.33$^{\ast}$\\
\bottomrule
\end{tabular}
\caption{\textbf{Eleven public ImageNet-1k checkpoints read at sixteen times their training token count} (896~px; 1{,}024~px
for the register models), all 50{,}000 validation images. The rescaling multiplies every score by $\ln N/\ln\Ntr$ at test
time. Share recovered: (rescaled $-$ as trained)/(training-resolution accuracy $-$ as trained). $^{\ast}$Median over
the eleven. The other rungs are in the supplement.}
\label{tab:checkpoints}
\end{table*}

\paragraph{Eleven public checkpoints.} Read at sixteen times their training token count, the eleven classifiers lose much
of their accuracy, and multiplying every score by $\lam$ at test time, with the weights unchanged, recovers part of it
(\cref{tab:checkpoints}, \cref{fig:teaser}c). The gain is positive in 10{} of the 11{} on enlarged crops,
by up to 21.9{} points, and in all 11{} on resized originals, by 0.6{} to 18.8{} points.
Pooled over the six primary models it is 7.68{} points on crops ([2.10, 14.31]) and 6.94{} on
originals ([2.51, 12.44]). It is at least as large with rotary positions (12.40, [10.46, 13.61]), so it is
not an artifact of interpolating absolute position embeddings, and it holds in the register models with average pooling
(3.69, [3.12, 4.25]). It grows with the token count: from 8 to 16 times the tokens it rises in
6{} of 6{} primary models, by 4.61{} points pooled ([1.85, 7.90]). Across the
eleven, the models that lose more at 16 times the tokens also gain more from the rescaling (Spearman's $\rho=0.57$
on crops, 0.64{} on originals).

\paragraph{What it does not recover.} At the largest evaluated resolution, the rescaling recovers a median of
0.23{} of the lost accuracy on crops (\textminus{}0.02{} to 0.45) and 0.33{} on originals (0.05{} to
0.49), and no public classifier returns to its training-resolution accuracy. On resized originals the two most
accurate checkpoints recover the least. AugReg2 ViT-B loses accuracy on enlarged crops at 8 and 16 times the tokens
(\textminus{}2.22{} and \textminus{}0.76{} points) and on originals at 4 times (\textminus{}0.22). On originals its estimate at 8 times
(\textminus{}0.28) has an interval spanning zero; it gains at 16 times (+2.38);
\cref{sec:multiplier} returns to it.

\paragraph{Real detail.} The native-detail subset keeps the 448{} validation images, from 196{}
classes, whose originals carry full detail at 896~px. On their resized originals the pooled gain is +5.7{}
points ([0.4, 12.0]). At 640~px, on 1{,}019{} images, the interval spans zero
(+1.7, [\textminus{}0.5, 4.4]). The rotary models gain 8.9{} to 12.9{} points
on this subset at 896~px, with intervals above zero in 3{} of 3. Real detail raises
accuracy before rescaling. As for the rescaling's gain, at 640~px real detail leaves it unchanged to within 1.5 points
(difference +0.13, [\textminus{}1.23, 1.44]), and at 896~px the subset is too small
to decide (+0.07, [\textminus{}2.72, 3.01]).

\paragraph{At twice the tokens.} At twice the tokens, on resized originals, the rescaled models exceed
their training-resolution accuracy in 7{} of 11{} (paired intervals above zero). Much of that is the
familiar benefit of testing above the training resolution~\citep{touvron2019fixres}, which 7{} models
already show without rescaling; the rescaling keeps and adds to it.

\paragraph{The gain needs no sink.} In the controlled models at eight times the tokens, every attention rule gains alike:
plain softmax 4.49{} points, zero-value registers 4.08, the learned sink
3.45, the gate 3.77{} and the gated sink 3.27{} (three-seed means on held-out
images), and moving the sink inside or outside the multiplier changes accuracy by at most 0.04{} points. A separate
test found the same at sixteen times the tokens: plain softmax gained 4.98{} points and the
learned sink 4.87~\citep{bruce2026sinks}. Accuracy therefore tests the growing score range, which every rule
shares; the registers' shares (\cref{sec:dilution}) test the sink. At sixteen times the tokens every controlled rule
gains 5.13{} to 5.76{} points on held-out images. At the same token count, the registers
gain 5.30{} to 6.82{} points and the gate 4.60{} to 5.36{} on all 5{,}000 images
in every seed.
\section{Which multiplier}\label{sec:multiplier}

\begin{figure}[t]
\centering
\includegraphics{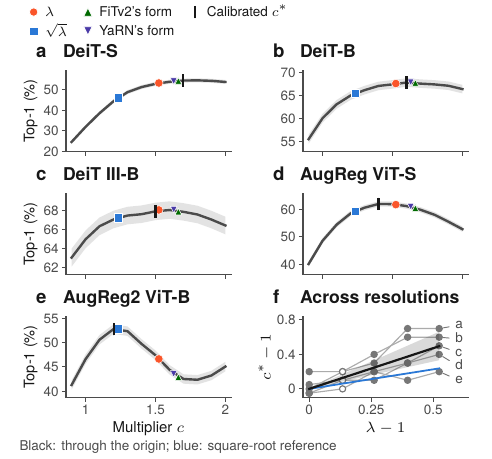}
\caption{\textbf{The best multiplier follows the log ratio on average.}
(a--e) Held-out top-1 on enlarged crops at 896~px (16$\times$ patches); black ticks are optima chosen on separate calibration images.
Shading: pointwise 95\% image-bootstrap intervals; vertical scales differ. AugReg2 ViT-B favours $\sqrt{\lam}$.
(f) Optima across resolutions; letters identify models. Hollow points are 320~px, excluded from every fit.
Black line and band: slope through the origin, 0.945{} [0.635, 1.266]; blue: square-root reference,
0.454. Slope 1 corresponds to $\lam$. Intervals are 95\% hierarchical bootstrap. All fits use the three
largest enlarged resolutions; the omitted free-intercept and measured-optimum fits (0.882, 0.875) also use
the measured training-resolution optimum.}
\label{fig:multiplier}
\end{figure}

\paragraph{The rules compared.} All comparisons use the enlarged-crop route. With $L=\ln(N/\Ntr)$, we compare the log ratio $\lam$ with the rules in use:
\begin{itemize}[leftmargin=*,itemsep=0pt]
\item $c=1$;
\item the square-root rule $\sqrt{\lam}$, used for image generators~\citep{jin2023training,gao2024lumina};
\item FiTv2's form $\max(1,\sqrt{L})$~\citep{wang2024fitv2};
\item YaRN's temperature transposed from context length to token count, $(0.1L+1)^2$~\citep{peng2024yarn};
\item a single multiplier $c^*$, calibrated separately for each model and evaluation resolution on the labelled
calibration half as the best value on a grid (0.9 to 2.0 in steps of 0.1 for the public models, 0.05 for the controlled
ones).
\end{itemize} For the public models the halves are the first 20 validation images of each class, ten
for calibration and ten for evaluation; for the controlled models, alternate images within each class. Every rule keeps
the registers inside the multiplier, and the output gate is never scaled. All are scored on the 10{,}000 held-out
images of the five primary models with absolute positions, at 4, 8 and 16 times the training token count.

\begin{table*}[t]
\centering
\small
\setlength{\tabcolsep}{6pt}
\begin{tabular}{@{}lcccccc@{}}
\toprule
Model & $c=1$ & $\sqrt{\lam}$ & $\lam$ & FiTv2 & YaRN & $c^*$ (value)\\
\midrule
DeiT III-B & 65.0 & 67.2 & 68.1 & 67.9 & 68.1 & 67.9 (1.5)\\
DeiT-B & 60.1 & 65.4 & 67.5 & 67.7 & 67.8 & 67.8 (1.6)\\
DeiT-S & 31.8 & 46.1 & 53.2 & 54.2 & 54.0 & 54.4 (1.7)\\
AugReg2 ViT-B & 46.6 & 52.8 & 46.6 & 42.9 & 43.5 & 52.8 (1.2)\\
AugReg ViT-S & 48.6 & 59.4 & 61.7 & 60.5 & 60.9 & 62.0 (1.4)\\
\bottomrule
\end{tabular}
\caption{\textbf{Multiplier rules on held-out enlarged crops} at 16$\times$ tokens (896~px), top-1 on 10{,}000 images not used to
choose $c^*$. Each $c^*$ is calibrated for this model at this resolution. Pooled over the five models and measured from
$\lam$, FiTv2's form is \textminus{}0.81{} points, YaRN's form \textminus{}0.55{} and $c^*$ +1.53{}
(Holm-adjusted $p=0.168$). The supplement gives every comparison's pointwise and simultaneous intervals.}
\label{tab:rules}
\end{table*}

\paragraph{On average, the log ratio.} The calibrated multiplier rises with $\lam$: across the five models the slope of
$c^*-1$ on $\lam-1$ is 0.945{} ([0.635, 1.266]), which excludes the square-root rule's 0.454{}
(\cref{fig:multiplier}). Model by model the slope runs from 0.39{} to 1.45. On held-out images,
$\lam$ comes within a point of the calibrated optimum in 3{} of 5{} models. FiTv2's and YaRN's forms
have lower point estimates than $\lam$, but neither difference is resolved (\textminus{}0.81{} and \textminus{}0.55{}
points). The log ratio also beats $\sqrt{\lam}$ in 4{} of five models, although the panel's difference,
+1.24{} points ([\textminus{}2.87, 4.87]), is not resolved (\cref{tab:rules}). In the fifteen
controlled checkpoints the picture is sharper. There $\lam$ beats $\sqrt{\lam}$ in 60{} of
60{} comparisons, by +1.85{} points at 16 times the tokens
([1.50, 2.20]), and it comes within a point of the calibrated optimum in
15{} of 15.

\paragraph{At the training resolution.} Every named rule equals one at 224~px, so calibrating the multiplier there tests
whether sharpening already helps at the training resolution. The calibrated multiplier lies between 0.95{} and
1.20{} in every model. It is within $1\pm0.1$ in 4{} of 5{} public models and 12{}
of 15{} controlled checkpoints. Choosing it gives no resolved held-out accuracy gain: \textminus{}0.022{} points
[\textminus{}0.204, 0.084] for the public models and +0.027{} [\textminus{}0.219, 0.144] for the controlled
checkpoints. The curves are flat near one, so the location of the optimum moves more than the
accuracy it buys. In the public models the fits with a measured anchor agree with the one through the origin, and all
exclude the square-root rule (supplement). The controlled optimum
is not linear either: its mean is 1.06, 1.34, 1.53, 1.60{} and
1.58{} at 1, 2, 4, 8 and 16 times the tokens, so no single slope fits every model.

\paragraph{A multiplier fixed at the training size leaves most of the gain unrealized.} We ask whether the models
simply want
sharper attention at any size. If they did, the multiplier calibrated at 224~px would carry the gain to higher
resolutions. Held fixed in each controlled checkpoint, this calibrated multiplier does help a little at higher
resolutions (+1.16{} points over $c=1$ at 16 times the tokens, [0.39,
1.98]). But $\lam$ does far better: by +2.93{} points at
8 times ([2.11, 3.58]) and +4.17{} at 16 times ([3.29,
5.00]), with the interval above zero in 15{} of 15 checkpoints. The intervals redo the
calibration in every resample.

\paragraph{Where a calibrated constant helps.} The pooled estimate for a constant calibrated per model is
+1.53{} points above $\lam$ (Holm-adjusted $p=0.168$; simultaneous interval [\textminus{}0.06, 4.92]);
the pooled difference remains unresolved. Almost all of this point-estimate advantage comes from AugReg2 ViT-B, where
the constant recovers +6.15{} points over $\lam$. For this model, $c^*=1.2$ is close to $\sqrt{\lam}$ at
this rung, against $\lam=1.524$. The controlled models lean the other way: at about twice the tokens, where
$\lam=1.135$, each of the 15{} checkpoints chooses $c^*$ between 1.20{}
and 1.50{} (median 1.35), whereas the public models choose 1.0{} to
1.2. \Cref{eq:keep} reads both: decisive heads want less than $\lam$, and less decisive heads want more.
For each recorded image, layer and head, we measure the top patch's score lead over the log-mean-exp of the other
patch scores, then take the median over these observations for each model. This equals $g$ in \cref{eq:keep} when the
other scores are equal; for unequal scores it is a descriptive summary. At the training size AugReg2 ViT-B is the most
decisive of the six primary models, with a median margin of 2.52{} against 1.75{} to
2.31{} for the rest. Across the five models with a calibrated multiplier, the median
margin is inversely associated with the selected multiplier at the largest resolution (Spearman's
$\rho={}$\textminus{}0.70). The idealized calculation agrees with the direction of this association, but does not reproduce
the exact ranking or the multiplier values: applied to the median margin, it asks for 2.10{} to
2.59, where the models choose 1.2{} to 1.7. These measurements do not identify which
heads determine the accuracy-optimal multiplier. In ViTs trained from scratch, a constant fitted just above the training size and carried
higher matched $\lam$ (mean difference 0.01{} points)~\citep{bruce2026sinks}.
\section{Limits and discussion}\label{sec:discussion}

\paragraph{What the rescaling does not do.} It does not restore the attention a model learned at its training size:
in the controlled models it moves the class token's attention further from its 224~px form (total variation up by
0.027{} to 0.045{} in 9 of 9{} checkpoints) and undoes only
17\%{} to 26\%{} of the entropy's rise~\citep{bruce2026sinks}. Why it helps beyond countering the dilution is open; it does not restore token diversity (supplement).

\paragraph{A second training-free correction.} Enlarging an image multiplies the patch keys by $m$ but not the class
token or the registers, so their share of attention falls. Adding $\ln m$ to the scores of these keys alone, in every
layer and head, restores that share exactly when each patch is repeated $m$ times. On held-out enlarged crops at
sixteen times the tokens, this offset gains +19.6{} points ([+18.7, +20.4]) in AugReg2 ViT-B, where
$\lam$ gains +0.04{} ([\textminus{}0.86, +0.93]), and +3.2{} to +16.4{} in three{} more of
the seven{} public models tested; it costs 1.5{} to 2.2{} points in the other three. With
$\lam$, ViT-Betwixt gains +14.4, against +3.0{} for $\lam$ alone. Its gains do not track
$\lam$'s (Spearman's $\rho={}$\textminus{}0.14), and in the controlled models it recovers none of $\lam$'s gain.
It helps the models that lose the most without rescaling, among only seven.

\paragraph{For practice.} For classification, these public models are more accurate at the training resolution than
at the largest tested resolution, even rescaled, though modest increases can help (\cref{sec:accuracy}). When a vision
transformer must read more tokens than it was trained on, without fine-tuning, multiply every attention score, the
registers' included, by $\ln N/\ln\Ntr$: it needs no training and no labels, and it is the identity at the training
size. Dense prediction and high-resolution encoders, where more tokens are the norm, are applications to test. Where a few thousand labelled images at the target size exist, calibrate one
constant instead.

\paragraph{Limits.} We study classification only, one checkpoint per public recipe and one controlled model size. Most
inputs are enlarged or resized; the native-detail subset is small. The enlarged-crop route holds image detail fixed
while changing patch sampling and positional representations along with token count. Production models partly compensate for the dilution,
and trained heads need not meet the law's margins, so its guarantee is a statement about an idealized setting. The law
counts keys, not pixels, so it also bears on a language model's context length~\citep{bruce2026sinks}.

\paragraph{Conclusion.} More patches thin a vision transformer's attention as a law of softmax attention predicts, and
multiplying every score by $\ln N/\ln\Ntr$ at test time recovers part of the lost accuracy without training. The
calibrated multiplier follows the log ratio on average, and in the public models larger median margins at the training
size go with smaller multipliers.
\section*{LLM Usage Statement}
AI tools were used, under the author's direction, for several kinds of work:
software development and debugging, mathematical derivation and checking, data analysis, literature search and citation
checking, and drafting and editing of the text. The author reviewed and verified all AI-assisted content and
takes full responsibility for the paper.
{
    \small
    \bibliographystyle{ieeenat_fullname}
    \bibliography{main}

\begin{thebibliography}{34}
\providecommand{\natexlab}[1]{#1}
\providecommand{\url}[1]{\texttt{#1}}
\expandafter\ifx\csname urlstyle\endcsname\relax
  \providecommand{\doi}[1]{doi: #1}\else
  \providecommand{\doi}{doi: \begingroup \urlstyle{rm}\Url}\fi

\bibitem[Bai et~al.(2023)]{bai2023qwen}
Jinze Bai et~al.
\newblock Qwen technical report.
\newblock arXiv preprint arXiv:2309.16609, 2023.

\bibitem[Beyer et~al.(2023)Beyer, Izmailov, Kolesnikov, Caron, Kornblith, Zhai,
  Minderer, Tschannen, Alabdulmohsin, and Pavetic]{beyer2023flexivit}
Lucas Beyer, Pavel Izmailov, Alexander Kolesnikov, Mathilde Caron, Simon
  Kornblith, Xiaohua Zhai, Matthias Minderer, Michael Tschannen, Ibrahim
  Alabdulmohsin, and Filip Pavetic.
\newblock {FlexiViT}: One model for all patch sizes.
\newblock In \emph{IEEE/CVF Conference on Computer Vision and Pattern
  Recognition (CVPR)}, pages 14496--14506, 2023.

\bibitem[Bruce(2026)]{bruce2026sinks}
Neil D.~B. Bruce.
\newblock Attention sinks at long context: One log-{N} law for retrieval and
  abstention.
\newblock Technical report, Zenodo, 2026.
\newblock \url{https://doi.org/10.5281/zenodo.23541963}.

\bibitem[Chiang and Cholak(2022)]{chiang2022overcoming}
David Chiang and Peter Cholak.
\newblock Overcoming a theoretical limitation of self-attention.
\newblock In \emph{Proceedings of the 60th Annual Meeting of the Association
  for Computational Linguistics (Volume 1: Long Papers)}, pages 7654--7664.
  Association for Computational Linguistics, 2022.

\bibitem[Darcet et~al.(2024)Darcet, Oquab, Mairal, and
  Bojanowski]{darcet2024registers}
Timoth{\'e}e Darcet, Maxime Oquab, Julien Mairal, and Piotr Bojanowski.
\newblock Vision {Transformers} need registers.
\newblock In \emph{International Conference on Learning Representations}, 2024.

\bibitem[Dehghani et~al.(2023)Dehghani, Mustafa, Djolonga, Heek, Minderer,
  Caron, Steiner, Puigcerver, Geirhos, Alabdulmohsin, Oliver, Padlewski,
  Gritsenko, Lu{\v{c}}i{\'c}, and Houlsby]{dehghani2023navit}
Mostafa Dehghani, Basil Mustafa, Josip Djolonga, Jonathan Heek, Matthias
  Minderer, Mathilde Caron, Andreas Steiner, Joan Puigcerver, Robert Geirhos,
  Ibrahim Alabdulmohsin, Avital Oliver, Piotr Padlewski, Alexey Gritsenko,
  Mario Lu{\v{c}}i{\'c}, and Neil Houlsby.
\newblock Patch n' pack: {NaViT}, a vision transformer for any aspect ratio and
  resolution.
\newblock In \emph{Advances in Neural Information Processing Systems}, pages
  2252--2274, 2023.

\bibitem[Deng et~al.(2009)Deng, Dong, Socher, Li, Li, and
  Fei-Fei]{deng2009imagenet}
Jia Deng, Wei Dong, Richard Socher, Li-Jia Li, Kai Li, and Li Fei-Fei.
\newblock {ImageNet}: A large-scale hierarchical image database.
\newblock In \emph{IEEE Conference on Computer Vision and Pattern Recognition
  (CVPR)}, pages 248--255, 2009.

\bibitem[Dosovitskiy et~al.(2021)Dosovitskiy, Beyer, Kolesnikov, Weissenborn,
  Zhai, Unterthiner, Dehghani, Minderer, Heigold, Gelly, Uszkoreit, and
  Houlsby]{dosovitskiy2021vit}
Alexey Dosovitskiy, Lucas Beyer, Alexander Kolesnikov, Dirk Weissenborn,
  Xiaohua Zhai, Thomas Unterthiner, Mostafa Dehghani, Matthias Minderer, Georg
  Heigold, Sylvain Gelly, Jakob Uszkoreit, and Neil Houlsby.
\newblock An image is worth 16x16 words: {Transformers} for image recognition
  at scale.
\newblock In \emph{International Conference on Learning Representations}, 2021.

\bibitem[Fuller et~al.(2024)Fuller, Kyrollos, Yassin, and
  Green]{fuller2024lookhere}
Anthony Fuller, Daniel Kyrollos, Yousef Yassin, and James Green.
\newblock {LookHere}: Vision transformers with directed attention generalize
  and extrapolate.
\newblock In \emph{Advances in Neural Information Processing Systems}, pages
  19683--19739, 2024.

\bibitem[Gao et~al.(2024)]{gao2024lumina}
Peng Gao et~al.
\newblock {Lumina-T2X}: Transforming text into any modality, resolution, and
  duration via flow-based large diffusion transformers.
\newblock arXiv preprint arXiv:2405.05945, 2024.

\bibitem[Heo et~al.(2024)Heo, Park, Han, and Yun]{heo2024rotary}
Byeongho Heo, Song Park, Dongyoon Han, and Sangdoo Yun.
\newblock Rotary position embedding for vision transformer.
\newblock In \emph{European Conference on Computer Vision (ECCV)}, pages
  289--305, 2024.

\bibitem[Jiang et~al.(2025)Jiang, Dravid, Efros, and
  Gandelsman]{jiang2025registers}
Nicholas Jiang, Amil Dravid, Alexei Efros, and Yossi Gandelsman.
\newblock Vision transformers don't need trained registers.
\newblock In \emph{Advances in Neural Information Processing Systems}, pages
  56557--56595, 2025.

\bibitem[Jin et~al.(2023)Jin, Shen, Li, and Xue]{jin2023training}
Zhiyu Jin, Xuli Shen, Bin Li, and Xiangyang Xue.
\newblock Training-free diffusion model adaptation for variable-sized
  text-to-image synthesis.
\newblock In \emph{Advances in Neural Information Processing Systems}, pages
  70847--70860, 2023.

\bibitem[Leroy et~al.(2024)Leroy, Revaud, Lucas, and
  Weinzaepfel]{leroy2024winwin}
Vincent Leroy, Jerome Revaud, Thomas Lucas, and Philippe Weinzaepfel.
\newblock {Win-Win}: Training high-resolution vision transformers from two
  windows.
\newblock In \emph{International Conference on Learning Representations}, 2024.

\bibitem[Miller(2023)]{miller2023offbyone}
Evan Miller.
\newblock Attention is off by one.
\newblock Blog post,
  \url{https://www.evanmiller.org/attention-is-off-by-one.html}, 2023.

\bibitem[Nakanishi(2025)]{nakanishi2025ssmax}
Ken~M. Nakanishi.
\newblock {Scalable-Softmax} is superior for attention.
\newblock arXiv preprint arXiv:2501.19399, 2025.

\bibitem[{OpenAI}(2025)]{openai2025gptoss}
{OpenAI}.
\newblock {gpt-oss-120b} \& {gpt-oss-20b} model card.
\newblock Model card, \url{https://openai.com/index/gpt-oss-model-card/}, 2025.
\newblock Released 5 August 2025; also arXiv:2508.10925.

\bibitem[Oquab et~al.(2024)Oquab, Darcet, Moutakanni, Vo, Szafraniec, Khalidov,
  Fernandez, Haziza, Massa, El-Nouby, Assran, Ballas, Galuba, Howes, Huang, Li,
  Misra, Rabbat, Sharma, Synnaeve, Xu, J{\'e}gou, Mairal, Labatut, Joulin, and
  Bojanowski]{oquab2024dinov2}
Maxime Oquab, Timoth{\'e}e Darcet, Th{\'e}o Moutakanni, Huy~V. Vo, Marc
  Szafraniec, Vasil Khalidov, Pierre Fernandez, Daniel Haziza, Francisco Massa,
  Alaaeldin El-Nouby, Mido Assran, Nicolas Ballas, Wojciech Galuba, Russell
  Howes, Po-Yao Huang, Shang-Wen Li, Ishan Misra, Michael Rabbat, Vasu Sharma,
  Gabriel Synnaeve, Hu Xu, Herv{\'e} J{\'e}gou, Julien Mairal, Patrick Labatut,
  Armand Joulin, and Piotr Bojanowski.
\newblock {DINOv2}: Learning robust visual features without supervision.
\newblock \emph{Transactions on Machine Learning Research}, 2024.

\bibitem[Peng et~al.(2024)Peng, Quesnelle, Fan, and Shippole]{peng2024yarn}
Bowen Peng, Jeffrey Quesnelle, Honglu Fan, and Enrico Shippole.
\newblock {YaRN}: Efficient context window extension of large language models.
\newblock In \emph{International Conference on Learning Representations}, 2024.

\bibitem[Qiu et~al.(2025)]{qiu2025gated}
Zihan Qiu et~al.
\newblock Gated attention for large language models: Non-linearity, sparsity,
  and attention-sink-free.
\newblock In \emph{Advances in Neural Information Processing Systems}, pages
  100092--100118. Curran Associates, Inc., 2025.

\bibitem[Russakovsky et~al.(2015)Russakovsky, Deng, Su, Krause, Satheesh, Ma,
  Huang, Karpathy, Khosla, Bernstein, Berg, and Fei-Fei]{russakovsky2015ilsvrc}
Olga Russakovsky, Jia Deng, Hao Su, Jonathan Krause, Sanjeev Satheesh, Sean Ma,
  Zhiheng Huang, Andrej Karpathy, Aditya Khosla, Michael Bernstein,
  Alexander~C. Berg, and Li Fei-Fei.
\newblock {ImageNet} large scale visual recognition challenge.
\newblock \emph{International Journal of Computer Vision}, 115:\penalty0
  211--252, 2015.

\bibitem[Steiner et~al.(2022)Steiner, Kolesnikov, Zhai, Wightman, Uszkoreit,
  and Beyer]{steiner2022augreg}
Andreas~Peter Steiner, Alexander Kolesnikov, Xiaohua Zhai, Ross Wightman, Jakob
  Uszkoreit, and Lucas Beyer.
\newblock How to train your {ViT}? {D}ata, augmentation, and regularization in
  vision transformers.
\newblock \emph{Transactions on Machine Learning Research}, 2022.

\bibitem[Su(2021)]{su2021entropy}
Jianlin Su.
\newblock Viewing the scale operation of attention from the perspective of
  entropy invariance.
\newblock Scientific Spaces (blog), \url{https://kexue.fm/archives/8823}, 2021.
\newblock In Chinese; English translation of the original title.

\bibitem[Sun et~al.(2024)Sun, Chen, Kolter, and Liu]{sun2024massive}
Mingjie Sun, Xinlei Chen, J.~Zico Kolter, and Zhuang Liu.
\newblock Massive activations in large language models.
\newblock In \emph{Conference on Language Modeling}, 2024.

\bibitem[Tian et~al.(2020)Tian, Krishnan, and Isola]{tian2020cmc}
Yonglong Tian, Dilip Krishnan, and Phillip Isola.
\newblock Contrastive multiview coding.
\newblock In \emph{European Conference on Computer Vision (ECCV)}, pages
  776--794, 2020.

\bibitem[Touvron et~al.(2019)Touvron, Vedaldi, Douze, and
  J{\'e}gou]{touvron2019fixres}
Hugo Touvron, Andrea Vedaldi, Matthijs Douze, and Herv{\'e} J{\'e}gou.
\newblock Fixing the train-test resolution discrepancy.
\newblock In \emph{Advances in Neural Information Processing Systems}, 2019.

\bibitem[Touvron et~al.(2021)Touvron, Cord, Douze, Massa, Sablayrolles, and
  J{\'e}gou]{touvron2021training}
Hugo Touvron, Matthieu Cord, Matthijs Douze, Francisco Massa, Alexandre
  Sablayrolles, and Herv{\'e} J{\'e}gou.
\newblock Training data-efficient image transformers \& distillation through
  attention.
\newblock In \emph{International Conference on Machine Learning (ICML)}, pages
  10347--10357, 2021.

\bibitem[Touvron et~al.(2022)Touvron, Cord, and J{\'e}gou]{touvron2022deit3}
Hugo Touvron, Matthieu Cord, and Herv{\'e} J{\'e}gou.
\newblock {DeiT III}: Revenge of the {ViT}.
\newblock In \emph{European Conference on Computer Vision (ECCV)}, pages
  516--533, 2022.

\bibitem[Veli{\v{c}}kovi{\'c} et~al.(2025)Veli{\v{c}}kovi{\'c},
  Perivolaropoulos, Barbero, and Pascanu]{velickovic2025softmax}
Petar Veli{\v{c}}kovi{\'c}, Christos Perivolaropoulos, Federico Barbero, and
  Razvan Pascanu.
\newblock Softmax is not enough (for sharp size generalisation).
\newblock In \emph{Proceedings of the 42nd International Conference on Machine
  Learning}, 2025.

\bibitem[Wang(2026)]{wang2026abstention}
Richard~Zhe Wang.
\newblock Abstention and noise filtering: Two missing primitives of softmax
  attention.
\newblock arXiv preprint arXiv:2609.22005, 2026.

\bibitem[Wang et~al.(2024)Wang, Lu, Huang, Zhou, Yue, Ouyang, and
  Bai]{wang2024fitv2}
ZiDong Wang, Zeyu Lu, Di Huang, Cai Zhou, Xiaoyu Yue, Wanli Ouyang, and Lei
  Bai.
\newblock {FiTv2}: Scalable and improved flexible vision transformer for
  diffusion model.
\newblock arXiv preprint arXiv:2410.13925, 2024.

\bibitem[Wightman(2019)]{wightman2019timm}
Ross Wightman.
\newblock {PyTorch} image models.
\newblock GitHub repository, 2019.

\bibitem[Xiao et~al.(2024)Xiao, Tian, Chen, Han, and
  Lewis]{xiao2024streamingllm}
Guangxuan Xiao, Yuandong Tian, Beidi Chen, Song Han, and Mike Lewis.
\newblock Efficient streaming language models with attention sinks.
\newblock In \emph{International Conference on Learning Representations}, 2024.

\bibitem[Yan et~al.(2026)Yan, Liu, Wang, Lu, Zhang, Li, and
  Ji]{yan2026extravar}
Feihong Yan, Shaoyu Liu, Haixuan Wang, Shuai Lu, Linfeng Zhang, Huiqi Li, and
  Xiangyang Ji.
\newblock {ExtraVAR}: Stage-aware {RoPE} remapping for resolution extrapolation
  in visual autoregressive models.
\newblock arXiv:2605.10045, 2026.

\end{thebibliography}
}

\clearpage
\setcounter{section}{0}\renewcommand{\theHsection}{supp.\arabic{section}}
\maketitlesupplementary
\renewcommand{\thetable}{S\arabic{table}}\setcounter{table}{0}
\renewcommand{\thefigure}{S\arabic{figure}}\setcounter{figure}{0}
\renewcommand{\thesection}{S\arabic{section}}

\section{The cited results, restated}\label{sup:cited}
The main paper uses the following sink and rescaling results of Bruce~\citep{bruce2026sinks}, with their original
numbering, and the dilution consequences and extension below. Take one attention row with $N$ content keys of scores
$s_j$ and a sink of score $b$, so that the sink's share is $a_N=e^b/(e^b+B_N)$ with $B_N=\sum_je^{s_j}$. Counts are
positive integers. The rescaling statements below use $N\ge\Ntr>1$, and the targets satisfy $a_*,w_*\in(0,1)$, with
$A_*=a_*/(1-a_*)$ and $W_*=w_*/(1-w_*)$.

\paragraph{Dilution (from the concentration argument of Prop.~7).} Suppose the exponentiated scores $e^{s_j}$ are
independent, with a common positive mean $\mu$ and a common upper bound $u$. Then $B_N/N\to\mu$ almost surely: Hoeffding's
inequality bounds $\mathbb{P}(|B_N/N-\mu|\ge t)$ by $2e^{-2Nt^2/u^2}$, which is summable in $N$, so the Borel--Cantelli lemma
applies. Three consequences follow. The crowd's log-sum is $\ln B_N=\ln N+\ln\mu+o(1)$. A sink of fixed score $b$ has a
share that concentrates near $e^b/(e^b+N\mu)$, and its log-odds $b-\ln B_N$ plus $\ln N$ tend to $b-\ln\mu$, so they fall
like $-\ln N$. And the row's entropy grows like $\ln N$, while the top score, being bounded, does not grow. For the entropy,
write $X_j=e^{s_j}\in(0,u]$ and $Z_N=e^b+\sum_jX_j$. Then
\[
H_N=\ln Z_N-\frac{b\,e^b+\sum_jX_j\ln X_j}{Z_N}.
\]
Since $x\ln x$ is bounded on $(0,u]$, the numerator of the fraction is $O(N)$, while $Z_N\ge B_N$ grows like $N\mu$; so the
fraction stays bounded, and $H_N=\ln N+O(1)$. For scores confined to a range that does not grow, independence is not
needed: the log-sum and the entropy still lie within a constant of $\ln N$.

\paragraph{Sinks are gates (Prop.~2).} A row with a sink factors as $P_{ij}=g_i\,\mathrm{softmax}_j(s_{ij})$ with
$g_i=\sigma(\LSE_js_{ij}-b_i)$, where $\sigma(x)=(1+e^{-x})^{-1}$. A zero-value register adds its exponentiated score to the sink and nothing to the output;
off-by-one softmax is the case $b_i=0$; a learned sink logit is a learned $b_i$. A headwise output gate multiplies the
head's output by a factor computed from the query, outside the softmax, so for a fixed query representation its share does not depend on $N$.

\paragraph{One sink cannot do two jobs (Thms.~4 and 8).} Let a query without a match need $a_N\ge a_*$ and a query with
a match of score $s_*$, among the same sink and crowd, need the match's weight $w_N\ge w_*$. Both hold if and only if
$A_*B_N\le e^b\le e^{s_*}/W_*-B_N$, with $A_*$ and $W_*$ the targets' odds. If every content score lies in
$[-\tau,\tau]$, a range that does not grow with $N$, no sink, fixed, learned or growing, meets both beyond
$N_*=e^{2\tau}(1-a_*)(1-w_*)/w_*$. Multiplying every score, the sink's included, by $\lam=\ln N/\ln\Ntr$ keeps both
under separated margins: if the crowd scores at most $s_0$ and $s_*>b>s_0$, then with $v_a=\Ntr e^{s_0-b}$,
$v_r=\Ntr e^{s_0-s_*}$ and $u_r=e^{b-s_*}$, the rescaled row has $a_N\ge(1+v_a^\lam)^{-1}$ and
$w_N\ge(1+u_r^\lam+v_r^\lam)^{-1}$. If $v_a\le1$ both bounds are non-decreasing in $N$, and if $v_a<1$ both tend to
one. Here $N$ and $\Ntr$ count content keys. If the multiplier is computed from total-key counts $T$ and $T_{\mathrm{tr}}$
instead, $\lambda_T=\ln T/\ln T_{\mathrm{tr}}$, then $e^{\lambda_Tx}=T^{x/\ln T_{\mathrm{tr}}}$, and since there are at most
$T$ content keys, the ratio of crowd to sink after rescaling is
\[
\sum_je^{\lambda_T(s_j-b)}\le Te^{\lambda_T(s_0-b)}=T^{\,1-(b-s_0)/\ln T_{\mathrm{tr}}} .
\]
It tends to zero when $b-s_0>\ln T_{\mathrm{tr}}$, and with $s_*>b$ the same two weights tend to one as before.

\paragraph{What the rescaling keeps (Thm.~9).} For $N=\Ntr^\lam$ and any finite scores, for $q>0$ write
$M_{q,N}=\bigl(N^{-1}\sum_je^{qs_j}\bigr)^{1/q}$. Then exactly
\[
 a_N=\Bigl[1+\bigl(\Ntr e^{-b}M_{\lam,N}\bigr)^\lam\Bigr]^{-1}.
\]
On a crowd that repeats the training scores $k$ times, for an integer $k$ with $N=k\Ntr$, $M_{\lam,N}=M_{\lam,\Ntr}$, so the rescaled share follows
from the scores at the training resolution alone. This is the prediction of (L4). For a group of registers with scores
$b_1,\dots,b_R$, the prediction replaces $b$ by $\LSE_rb_r$ and multiplies it by $\lam$; this is exact for $R=1$, and an
approximation otherwise, because $\LSE_r(\lam b_r)$ and $\lam\LSE_rb_r$ differ in general (they agree at $\lam=1$).

\subsection{Which multiplier preserves the top key's attention weight}\label{sup:keep}
Fix one head and one query, with $n_{\mathrm{tr}}\ge1$. At the training size, let its top key score $s_*$ and the
$n_{\mathrm{tr}}$ other keys score $s_1,\dots,s_{n_{\mathrm{tr}}}$. Read the head with $n=kn_{\mathrm{tr}}$ other keys,
for an integer $k\ge1$, whose scores repeat the training ones $k$ times. The log-ratio and square-root specializations
below use $n_{\mathrm{tr}}>1$. With every score multiplied by $c$, the top key's log-odds
against the others are
\[
 h(c)=c\,s_*-\LSE_j(c\,s_j)-L,\qquad L=\ln(n/n_{\mathrm{tr}})=\ln k,
\]
and its training log-odds are $h_0=s_*-\LSE_j(s_j)$. The multiplier that keeps the decision solves $h(c)=h_0$.

\begin{proposition}\label{prop:keep}
Let $s_*>\max_js_j$ and $L>0$. Write $g_1=s_*-\sum_jp_js_j$ for the top key's lead over the others' mean under the
training softmax $p_j=e^{s_j}/\sum_ke^{s_k}$, and $g_\infty=s_*-\max_js_j$ for its lead over the largest. Then $h(c)=h_0$
has exactly one solution $c_{\mathrm{keep}}>1$, and
\[
 1+\frac{L}{g_1}\ \le\ c_{\mathrm{keep}}\ \le\ 1+\frac{L}{g_\infty}.
\]
If the other keys score equally, $g_1=g_\infty=g$ and $c_{\mathrm{keep}}=1+L/g$ exactly, which is the main paper's \cref{eq:keep}. It
equals the log ratio $\ln n/\ln n_{\mathrm{tr}}$ when $g=\ln n_{\mathrm{tr}}$, and to first order in $L$ the square-root
rule keeps $g=2\ln n_{\mathrm{tr}}$. At $L=0$ the solution is $c=1$ for every head.
\end{proposition}

\begin{proof}
We study $h$ as a function of $c$. It is concave and increasing, so it has a unique root, and two lines through
$(1,h(1))$, its tangent and a line of slope $g_\infty$, bound that root from both sides. Write $p_j(c)=e^{cs_j}/\sum_ke^{cs_k}$ for the softmax of the scaled
other keys. Differentiating the log-sum-exp,
\[
h'(c)=s_*-\sum_jp_j(c)\,s_j,\qquad h''(c)=-\mathrm{Var}_{p(c)}(s)\le0,
\]
so $h$ is concave. A weighted mean of the $s_j$ is at most their maximum, so $h'(c)\ge g_\infty>0$ for every $c$.
Hence $h$ is strictly increasing, and $h(c)\ge h(1)+g_\infty(c-1)$ grows without bound. Since
$h(1)=h_0-L<h_0$, the equation $h(c)=h_0$ has exactly one root $c_{\mathrm{keep}}$, and $c_{\mathrm{keep}}>1$.

For the bounds, compare $h$ with two lines through $(1,h(1))$. A concave function lies below its tangent, and
$h'(1)=g_1$; the lower bound on $h'$ supplies the other line. For $c\ge1$,
\[
h(1)+g_\infty(c-1)\;\le\;h(c)\;\le\;h(1)+g_1(c-1).
\]
At $c=c_{\mathrm{keep}}$ we have $h(c)-h(1)=h_0-h(1)=L$, so $g_\infty(c_{\mathrm{keep}}-1)\le L\le g_1(c_{\mathrm{keep}}-1)$,
which are the two bounds.

If the other keys score equally, the variance vanishes, $h$ is linear and $g_1=g_\infty=g$, so
$c_{\mathrm{keep}}=1+L/g$ exactly. The log ratio $\lam=\ln n/\ln n_{\mathrm{tr}}=1+L/\ln n_{\mathrm{tr}}$ is this flat
case with $g=\ln n_{\mathrm{tr}}$. For the square-root rule,
\[
0\le1+\frac{L}{2\ln n_{\mathrm{tr}}}-\sqrt\lam\le\frac{L^2}{8\ln^2n_{\mathrm{tr}}},
\]
so to first order in $L$, $\sqrt\lam$ is the flat case with $g=2\ln n_{\mathrm{tr}}$. Finally, at $L=0$ we have
$h(1)=h_0$, so $c=1$ is the solution.
\end{proof}

\noindent For the same other keys, a head's decision is kept by a smaller multiplier the more its top key
leads, and for $L>0$ the bounds are tight only when its other keys score alike. The proposition uses the log ratio of the
other keys' counts. At the evaluated public-checkpoint token counts, the all-key multiplier differs from this ratio by
less than 0.0015. A numerical check solves for $c_{\mathrm{keep}}$ by bisection
on 3{,}000{} random heads and confirms both bounds, and the flat case to a relative error of 7e-16.

\section{The method in practice}\label{sup:method}
\paragraph{The change.} Multiply every attention score by $\lam$ before the softmax, in every layer and head. In a
timm vision transformer this is one assignment per attention module, since each module multiplies its scores by its
\texttt{scale}, $1/\sqrt{d}$:
{\footnotesize
\begin{verbatim}
import math
def rescale(model, res):
    p = model.patch_embed.patch_size
    k = model.num_prefix_tokens  # cls, regs
    n = (res // p[0]) * (res // p[1]) + k
    n0 = model.patch_embed.num_patches + k
    lam = math.log(n) / math.log(n0)
    for blk in model.blocks:
        a = blk.attn
        if not hasattr(a, "_resolution_base_scale"):
            a._resolution_base_scale = a.scale
        a.fused_attn = False
        a.scale = a._resolution_base_scale * lam
    return lam
\end{verbatim}}
The first call must receive an unmodified checkpoint; later calls assign from the cached base scale, so a resolution
sweep does not compound the multiplier. The model must accept the new size. In timm,
\texttt{dynamic\_img\_size=True} enables this by interpolating its position embeddings. With a fused attention kernel, pass the scale $\lam/\sqrt{d}$ to it explicitly instead.

\paragraph{How $N$ is counted.} In the public models $N$ counts every key: the patches, the class token and any
registers. $\Ntr$ counts the same keys at the size the position embeddings were trained for, for example 197 for a
ViT/16 at 224~px. In the controlled models $N$ counts the class token and the patches and leaves out the four zero-value
registers. At the training size $\lam=1$ and the rescaled model is bit-identical to the original,
which every run checks.

\paragraph{Cost.} None beyond the higher-resolution pass itself: one scalar per forward pass, no training and no labels.

\paragraph{Checks in every run.} Each attention module counts its calls, because a rescaling that never runs would
silently return the original model, and every run records those counts. The measured training-resolution accuracies of
all 11{} checkpoints agree with the published timm results to within 0.03{} points.

\section{Public checkpoints: protocol and full results}\label{sup:checkpoints}
\paragraph{Checkpoints and preprocessing.} \Cref{tab:sup-models} lists the eleven checkpoints. Each is loaded from timm
at a fixed revision whose weights are checked by SHA-256, and read with its own evaluation transform: resize the short
side to $\lfloor R/\text{crop}\rfloor$, take the central $R\times R$ crop, and normalize.

\paragraph{The two routes.} \emph{Enlarged crops} apply the transform at the training size and then upsample the
normalized tensor bicubically to $R$, so they add tokens but no content. \emph{Resized originals} apply the transform at
size $R$, so they carry whatever detail the original image has.

\paragraph{Full results.} \Cref{tab:sup-rungs-abs,tab:sup-rungs-other} give every rung and route: top-1 and top-5, as
trained and rescaled, with the paired gain and its 95\% bootstrap interval over all 50{,}000 images. The first rung
above the training size is about twice the tokens.

\section{Attention spread by layer}\label{sup:spread}
\Cref{tab:sup-spread} gives the class token's slopes per unit log patch count for the six primary models, as trained,
on enlarged crops from 224 to 896~px. For each image the row's statistics are taken in every head; medians are taken over images,
then heads, then layers. The crowd's slope is near 1 in the first third of the layers and lower in the last third, while
the entropy's stays near 1 throughout.

These slopes measure spread, not harm. Copying every key exactly $k$ times raises the crowd's log-sum and the row's
entropy by exactly $\ln k$, and leaves the top score unchanged: every term of the sum appears $k$ times, and each copy
takes $1/k$ of its original's weight, so each weight $p$ becomes $k$ weights $p/k$, which adds $\ln k$ to the entropy. Yet
for the same reason the row still gives each region of the image the same total weight, so the row's output is unchanged. Enlarged crops resemble this case, since
upsampling adds no content, which is why the accuracy measurements, not the slopes, test harm.

\section{Register, sink and gate shares}\label{sup:shares}
\paragraph{The shares.} The register mass is $a=\sum_{j\in R}p_j$, the attention on the register keys $R$; an inside
sink's share is its attention mass in the same way. An outside gate gives the head output $z=g\sum_jp_jv_j$, and its
plotted quantity is $a=1-g$, the suppressed fraction of the output. Every plotted coordinate is $\ln\bigl(a/(1-a)\bigr)$.

\paragraph{Which rows count.} The main paper (\cref{sec:setup}) gives the selection rule: a pair of a head and a
query row counts if its median share on the option over the bank images at the training resolution reaches the
selection threshold. \Cref{tab:sup-split} counts distinct selected layer, head and query-position combinations, once
each: only the class token's rows for DINOv2, and the recorded patch rows for the supervised register models. Images do
not multiply this count.

\paragraph{Clamping.} Shares are clamped to $[10^{-7},1-10^{-7}]$ before taking log-odds. 3{} of the
92{} cells clamp more than 1\% of their selected rows; they are flagged and kept.

\paragraph{Results.} \Cref{tab:sup-shares} gives every model's slopes, measured and predicted, as trained and rescaled.
Inside DINOv2's training range, from 224 to 336~px, the share barely falls (log-odds slopes \textminus{}0.08{} to
\textminus{}0.06), against \textminus{}0.50{} to \textminus{}0.35{} above 518~px. DINOv2 without registers improvises sinks
on a few high-norm patches.

\paragraph{The split.} Let $b_R=\LSE_rb_r$ be the log-sum-exp of the register scores and $\ell_{\mathrm{clip}}$ the
clamped register log-odds. We reconstruct $C_{\mathrm{rec}}=b_R-\ell_{\mathrm{clip}}$, which equals the remaining keys'
log-sum when clamping is inactive. Before aggregation the fitted slopes satisfy
$\ell_{\mathrm{clip}}'=b_R'-C_{\mathrm{rec}}'$ row by row. \Cref{tab:sup-split} reports separate medians of the fitted
slopes of these quantities, at three head thresholds, so the medians need not satisfy that equality; the effect of
clamping on those medians has not been quantified.

\section{Native detail}\label{sup:native}
\paragraph{Strata.} An image's detail at evaluated size $R$ is $d=\min\bigl(1,S/\lfloor R/\text{crop}\rfloor\bigr)$,
where $S$ is the original's short side. It is \emph{native} if $d=1$, \emph{partial} if $1/2\le d<1$, and \emph{low}
otherwise. The strata depend on image sizes alone. Native cells with fewer than 300 images
are reported but do not count toward a verdict.

\paragraph{Results.} \Cref{tab:sup-native} gives each model's gain on native-detail originals, and the difference
between its gains on originals and on crops of the same images. We call the gain unchanged when its interval lies inside $[-1.5,+1.5]$ points. Reweighting every class equally, as a check, changes the gain's sign in 11{}
of the 132{} model, rung and stratum cells.

\section{Multipliers in full}\label{sup:multipliers}
\paragraph{The rules.} All comparisons use the enlarged-crop route. With $L=\ln(N/\Ntr)$, the rules compared are: $c=1$; the square-root rule
$\sqrt{\lam}$~\citep{jin2023training,gao2024lumina}; $\lam$; FiTv2's form $\max(1,\sqrt{L})$~\citep{wang2024fitv2}; YaRN's
temperature transposed to token count, $(0.1L+1)^2$~\citep{peng2024yarn}; and the calibrated $c^*$, selected separately for each model and evaluation resolution as the best value on a
grid (0.9 to 2.0 in steps of 0.1 for the public models, 0.05 for the controlled ones) on the calibration half. For the
public models the halves are the first 20 validation images of each class, ten for calibration and ten for evaluation;
for the controlled models, alternate images within each class. Every rule keeps the registers inside the multiplier, and
the output gate is never scaled. At the training resolution (\cref{sec:multiplier}) every model is calibrated on a
finer grid, 0.70 to 2.00 in steps of 0.05.

\paragraph{The slope.} The slope of the calibrated multiplier is
$\kappa=\sum_R(c^*_R-1)(\lam_R-1)/\sum_R(\lam_R-1)^2$ over the enlarged rungs, a fit through the origin. The interval is
obtained by resampling models or checkpoints and calibration images, with the selection repeated in every resample.
Across models, the slope ranges from 0.39{} to 1.45. Measuring the training-resolution anchor instead of assuming it
changes the controlled slope: through the origin from the measured optimum it is 1.207{} ([1.081,
1.345]), and with a free intercept 1.032{} ([0.928, 1.182]), against 1.357{} with the
anchor assumed. In the public models the three fits agree (0.945, 0.875{} and 0.882), and all exclude the
square-root rule's 0.454.

\paragraph{The comparisons.} \Cref{tab:sup-rules} compares each rule with $\lam$ on the held-out images at 8 and 16
times the tokens. Within each group, the public models and the controlled checkpoints, the six comparisons with $\lam$
form one family: their simultaneous intervals are conservative 99.17\% Bonferroni percentile bootstrap intervals,
and their $p$-values are Holm-adjusted. These are two separate adjustments: an interval that excludes zero and a Holm
$p$ below 0.05 need not coincide. The comparison of $\lam$ with $\sqrt{\lam}$ is reported with its own pointwise 95\%
interval.

\section{Why the rescaling helps: three accounts tested}\label{sup:why}
\paragraph{The accounts.} We tested three accounts of why multiplying every score by $\lam$ recovers accuracy. All comparisons use enlarged crops at four times the training side (896~px; 1{,}024~px
for the 256~px models), on the 15{} controlled checkpoints and seven{}
public models (\cref{tab:sup-why}). Public-model accuracies use the evaluation half defined in \cref{sup:multipliers};
controlled-model accuracies use the full ImageNet-100 validation set.
\begin{itemize}
\item \emph{Multiplicity.} Enlargement copies the patches but not the class token, the registers or a sink. With
$m$ times as many patch keys as at the training size ($m=16$ here), the share of these uncopied keys falls. Adding $\ln m$ to their scores, and to
no others, restores that share exactly when every patch is repeated $m$ times, because it multiplies every
uncopied key's $e^{s_j}$ by $m$, as the repetition multiplies the patches' summed $e^{s_j}$ by $m$, so the factor
cancels when the weights are normalized. We apply this offset to real
enlargements, and to exact token replicas, in which each patch token is repeated $m$ times inside the model and nothing
else changes. A control adds the same offset to as many patch keys drawn at random. When combined with the rescaling,
the offset is added after multiplication: an uncopied key has score $\lam s_j+\ln m$, and a patch key has score
$\lam s_j$.
\item \emph{Token diversity.} After pooling the patch tokens to the training-size grid, the fraction of their total
squared norm remaining after subtracting their mean, measured in the last third of the layers.
\item \emph{Head type.} For each head at the training size, we renormalize each patch query's attention over patch
keys and measure the fraction in its $3\times3$ neighborhood. We take the median first over patch queries and then over
bank images. A head is local if this value is at least one half; we apply $\lam$ to the other heads only.
\end{itemize}

\paragraph{Controlled models.} None of the three accounts carries the gain (\cref{tab:sup-why}). The multiplier gains
5.2{} points (3.8{} to 6.8) against an untempered loss of 24.7. Exact replicas cost at most
0.9{} points, or 2.3{} with zero-value registers, so repetition alone is a small part of the loss. The
offset undoes the replicas' loss exactly, yet on enlargements it gains \textminus{}0.01{} points ([\textminus{}0.30,
+0.27]): a share of $\lam$'s gain of 0.00{} ([\textminus{}0.06, +0.05]). The random-key control gains nothing,
here or in the public models. Token diversity falls with resolution in every model, by 0.21{} to
0.33{} (out of at most 1), but the multiplier restores it in none of the 15{} controlled checkpoints.
It moves the final readout slightly towards its training-size form: the cosine to each model's training-resolution readout rises in
15{} of 15{} controlled checkpoints and six{} of the seven{} public models, by at most
0.21. The controlled models have almost no local heads (at most 1, in four{} checkpoints),
and limiting $\lam$ to the other heads changes its gain by \textminus{}0.17{} points ([\textminus{}0.43, +0.09]).

\paragraph{Public models.} Repetition matters more here: exact replicas cost 0.8{} to 6.1{} points.
The offset alone is a large correction in four{} of the seven{} models (+3.2{} to
+19.6{} points) and a small loss in the other three{} (\textminus{}2.2{} to \textminus{}1.5). AugReg2 ViT-B gains
+19.6{} points from the offset, where $\lam$ gains +0.04. With $\lam$ added, ViT-Betwixt gains +14.4,
against +3.0{} for $\lam$ alone. The models the offset helps are the four{} that lose the most without
rescaling (at least 31.5{} points, against at most 22.4); the pattern rests on seven{} models only. In the public models, $\lam$'s gain runs partly through local
heads: applying it to the other heads only changes the gain by \textminus{}4.4{} points ([\textminus{}8.3, \textminus{}1.9]) relative to applying it to every head.
In DeiT-S, $\lam$ on local heads alone gains +15.1, against +5.8{} on the other heads alone.

\paragraph{What remains open.} The rescaling's gain is not restored token diversity, and in the public models it does
not come from global heads alone. The offset is a separate correction: its gains do not track the multiplier's
(Spearman's $\rho={}$\textminus{}0.14{} across the public models), and in the controlled models it recovers none of the
multiplier's gain.

\section{Reproducibility}\label{sup:repro}
Every number and table is computed from the experiments' records; numbers quoted from Bruce~\citep{bruce2026sinks} are
taken from that work. The Lean~4 package that accompanies this paper as ancillary files
proves every mathematical statement and lists the declaration for each.

\clearpage
\begin{table*}[p]
\centering
\footnotesize
\begin{tabular}{@{}llrrrrl@{}}
\toprule
Model & timm tag & Params & Training px & $\Ntr$ & Crop & Interpolation\\
\midrule
\multicolumn{7}{@{}l}{\emph{Absolute positions, class token}}\\
DeiT-S & \texttt{deit\_small\_patch16\_224.fb\_in1k} & 22M & 224 & 197 & 0.9 & bicubic\\
DeiT-B & \texttt{deit\_base\_patch16\_224.fb\_in1k} & 87M & 224 & 197 & 0.9 & bicubic\\
DeiT III-B & \texttt{deit3\_base\_patch16\_224.fb\_in1k} & 87M & 224 & 197 & 0.9 & bicubic\\
AugReg ViT-S & \texttt{vit\_small\_patch16\_224.augreg\_in21k\_ft\_in1k} & 22M & 224 & 197 & 0.9 & bicubic\\
AugReg2 ViT-B & \texttt{vit\_base\_patch16\_224.augreg2\_in21k\_ft\_in1k} & 87M & 224 & 197 & 0.9 & bicubic\\
AugReg ViT-L & \texttt{vit\_large\_patch16\_224.augreg\_in21k\_ft\_in1k} & 304M & 224 & 197 & 0.9 & bicubic\\
\multicolumn{7}{@{}l}{\emph{Rotary positions}}\\
RoPE ViT-S & \texttt{vit\_small\_patch16\_rope\_224.naver\_in1k} & 22M & 224 & 197 & 0.9 & bicubic\\
RoPE ViT-B & \texttt{vit\_base\_patch16\_rope\_224.naver\_in1k} & 86M & 224 & 197 & 0.9 & bicubic\\
RoPE-Mixed ViT-B & \texttt{vit\_base\_patch16\_rope\_mixed\_224.naver\_in1k} & 86M & 224 & 197 & 0.9 & bicubic\\
\multicolumn{7}{@{}l}{\emph{Registers, average pooling}}\\
ViT-M, 4 registers & \texttt{vit\_medium\_patch16\_reg4\_gap\_256.sbb\_in1k} & 39M & 256 & 260 & 0.95 & bicubic\\
ViT-Betwixt, 4 registers & \texttt{vit\_betwixt\_patch16\_reg4\_gap\_256.sbb\_in1k} & 60M & 256 & 260 & 0.95 & bicubic\\
\bottomrule
\end{tabular}
\caption{\textbf{The eleven public checkpoints.} $\Ntr$ counts every key at the training size.}
\label{tab:sup-models}
\end{table*}

\begin{table*}[p]
\centering
\footnotesize
\begin{tabular}{@{}lrrlrrlrr@{}}
\toprule
& & Patches & & \multicolumn{2}{c}{Top-1} & & \multicolumn{2}{c}{Top-5}\\
\cmidrule(lr){5-6}\cmidrule(l){8-9}
Model & px & $\times$ & Route & Trained & Rescaled & Gain [95\% CI] & Trained & Rescaled\\
\midrule
\multicolumn{9}{@{}l}{\emph{Absolute positions, class token}}\\
DeiT-S & 224 & 1 & both & 79.9 & = & -- & -- & --\\
 & 320 & 2.0 & crops & 78.7 & 79.3 & +0.62 [0.47, 0.76] & 94.3 & 94.6\\
 & 320 & 2.0 & originals & 79.6 & 80.3 & +0.69 [0.54, 0.84] & 95.0 & 95.3\\
 & 448 & 4.0 & crops & 72.4 & 75.6 & +3.24 [3.01, 3.49] & 90.8 & 92.7\\
 & 448 & 4.0 & originals & 75.8 & 78.6 & +2.79 [2.57, 3.02] & 92.9 & 94.4\\
 & 640 & 8.2 & crops & 55.9 & 66.5 & +10.59 [10.24, 10.93] & 78.9 & 87.5\\
 & 640 & 8.2 & originals & 64.0 & 72.7 & +8.70 [8.38, 9.03] & 85.0 & 91.1\\
 & 896 & 16.0 & crops & 31.3 & 53.2 & +21.92 [21.52, 22.34] & 52.6 & 76.6\\
 & 896 & 16.0 & originals & 41.2 & 60.0 & +18.83 [18.43, 19.24] & 64.2 & 82.6\\
DeiT-B & 224 & 1 & both & 82.0 & = & -- & -- & --\\
 & 320 & 2.0 & crops & 80.9 & 81.2 & +0.32 [0.20, 0.45] & 95.0 & 95.2\\
 & 320 & 2.0 & originals & 82.0 & 82.5 & +0.45 [0.33, 0.56] & 95.8 & 95.9\\
 & 448 & 4.0 & crops & 77.5 & 78.7 & +1.13 [0.96, 1.31] & 93.3 & 93.8\\
 & 448 & 4.0 & originals & 79.8 & 80.9 & +1.12 [0.95, 1.29] & 94.6 & 95.0\\
 & 640 & 8.2 & crops & 70.0 & 73.6 & +3.59 [3.34, 3.83] & 89.0 & 91.1\\
 & 640 & 8.2 & originals & 74.3 & 77.0 & +2.72 [2.49, 2.96] & 91.4 & 92.9\\
 & 896 & 16.0 & crops & 59.7 & 67.1 & +7.40 [7.09, 7.71] & 81.7 & 87.0\\
 & 896 & 16.0 & originals & 63.3 & 69.6 & +6.25 [5.95, 6.55] & 84.4 & 88.6\\
DeiT III-B & 224 & 1 & both & 83.8 & = & -- & -- & --\\
 & 320 & 2.0 & crops & 82.4 & 82.6 & +0.14 [0.03, 0.26] & 95.9 & 96.0\\
 & 320 & 2.0 & originals & 83.5 & 83.6 & +0.11 [0.00, 0.22] & 96.5 & 96.6\\
 & 448 & 4.0 & crops & 78.6 & 79.3 & +0.66 [0.49, 0.84] & 94.1 & 94.6\\
 & 448 & 4.0 & originals & 81.2 & 81.6 & +0.41 [0.24, 0.57] & 95.5 & 95.7\\
 & 640 & 8.2 & crops & 72.5 & 73.9 & +1.45 [1.22, 1.69] & 90.5 & 91.8\\
 & 640 & 8.2 & originals & 76.1 & 77.2 & +1.05 [0.83, 1.27] & 92.7 & 93.5\\
 & 896 & 16.0 & crops & 64.9 & 67.8 & +2.96 [2.68, 3.24] & 85.1 & 88.1\\
 & 896 & 16.0 & originals & 67.0 & 69.8 & +2.81 [2.54, 3.09] & 86.9 & 89.3\\
AugReg ViT-S & 224 & 1 & both & 81.4 & = & -- & -- & --\\
 & 320 & 2.0 & crops & 80.6 & 80.7 & +0.11 [\textminus{}0.03, 0.26] & 95.7 & 95.8\\
 & 320 & 2.0 & originals & 81.7 & 82.0 & +0.26 [0.12, 0.41] & 96.3 & 96.5\\
 & 448 & 4.0 & crops & 76.5 & 77.6 & +1.17 [0.95, 1.39] & 94.0 & 94.7\\
 & 448 & 4.0 & originals & 79.0 & 80.3 & +1.25 [1.04, 1.46] & 95.0 & 95.6\\
 & 640 & 8.2 & crops & 66.9 & 71.8 & +4.92 [4.62, 5.22] & 88.5 & 91.6\\
 & 640 & 8.2 & originals & 70.7 & 74.8 & +4.19 [3.90, 4.48] & 90.5 & 93.1\\
 & 896 & 16.0 & crops & 48.5 & 61.4 & +12.90 [12.51, 13.29] & 73.0 & 84.8\\
 & 896 & 16.0 & originals & 54.5 & 65.2 & +10.78 [10.40, 11.16] & 78.6 & 87.5\\
AugReg2 ViT-B & 224 & 1 & both & 85.1 & = & -- & -- & --\\
 & 320 & 2.0 & crops & 84.4 & 84.3 & \textminus{}0.10 [\textminus{}0.23, 0.03] & 97.2 & 97.0\\
 & 320 & 2.0 & originals & 85.2 & 85.1 & \textminus{}0.11 [\textminus{}0.23, 0.01] & 97.5 & 97.4\\
 & 448 & 4.0 & crops & 80.0 & 80.2 & +0.26 [0.05, 0.47] & 95.5 & 95.2\\
 & 448 & 4.0 & originals & 82.7 & 82.5 & \textminus{}0.22 [\textminus{}0.41, \textminus{}0.02] & 96.4 & 96.1\\
 & 640 & 8.2 & crops & 68.4 & 66.1 & \textminus{}2.22 [\textminus{}2.55, \textminus{}1.90] & 88.8 & 86.1\\
 & 640 & 8.2 & originals & 73.2 & 72.9 & \textminus{}0.28 [\textminus{}0.58, 0.03] & 91.6 & 90.4\\
 & 896 & 16.0 & crops & 46.3 & 45.6 & \textminus{}0.76 [\textminus{}1.16, \textminus{}0.37] & 69.5 & 67.4\\
 & 896 & 16.0 & originals & 53.4 & 55.8 & +2.38 [2.00, 2.76] & 76.2 & 77.5\\
AugReg ViT-L & 224 & 1 & both & 85.8 & = & -- & -- & --\\
 & 320 & 2.0 & crops & 85.1 & 85.1 & \textminus{}0.01 [\textminus{}0.12, 0.10] & 97.6 & 97.6\\
 & 320 & 2.0 & originals & 86.0 & 86.1 & +0.05 [\textminus{}0.05, 0.16] & 97.9 & 97.9\\
 & 448 & 4.0 & crops & 82.9 & 82.6 & \textminus{}0.22 [\textminus{}0.39, \textminus{}0.04] & 96.7 & 96.8\\
 & 448 & 4.0 & originals & 84.5 & 84.7 & +0.19 [0.03, 0.34] & 97.4 & 97.5\\
 & 640 & 8.2 & crops & 78.8 & 78.9 & +0.10 [\textminus{}0.13, 0.32] & 94.9 & 95.4\\
 & 640 & 8.2 & originals & 80.7 & 81.0 & +0.29 [0.07, 0.50] & 95.8 & 96.3\\
 & 896 & 16.0 & crops & 72.0 & 73.7 & +1.67 [1.40, 1.94] & 90.9 & 92.5\\
 & 896 & 16.0 & originals & 74.6 & 75.2 & +0.60 [0.33, 0.88] & 92.4 & 93.5\\
\addlinespace
\bottomrule
\end{tabular}
\caption{\textbf{Every rung and route: absolute positions.} At the training size the rescaled model is the original.}
\label{tab:sup-rungs-abs}
\end{table*}

\begin{table*}[p]
\centering
\footnotesize
\begin{tabular}{@{}lrrlrrlrr@{}}
\toprule
& & Patches & & \multicolumn{2}{c}{Top-1} & & \multicolumn{2}{c}{Top-5}\\
\cmidrule(lr){5-6}\cmidrule(l){8-9}
Model & px & $\times$ & Route & Trained & Rescaled & Gain [95\% CI] & Trained & Rescaled\\
\midrule
\multicolumn{9}{@{}l}{\emph{Rotary positions}}\\
RoPE ViT-S & 224 & 1 & both & 81.2 & = & -- & -- & --\\
 & 320 & 2.0 & crops & 80.6 & 80.9 & +0.29 [0.15, 0.42] & 94.6 & 94.7\\
 & 320 & 2.0 & originals & 81.3 & 81.5 & +0.15 [0.02, 0.28] & 95.0 & 95.1\\
 & 448 & 4.0 & crops & 75.7 & 76.4 & +0.66 [0.45, 0.88] & 90.7 & 90.7\\
 & 448 & 4.0 & originals & 77.7 & 78.0 & +0.37 [0.17, 0.58] & 92.0 & 91.7\\
 & 640 & 8.2 & crops & 57.3 & 62.5 & +5.22 [4.91, 5.55] & 76.6 & 80.1\\
 & 640 & 8.2 & originals & 66.1 & 70.9 & +4.79 [4.50, 5.08] & 84.0 & 86.6\\
 & 896 & 16.0 & crops & 24.3 & 37.5 & +13.21 [12.84, 13.58] & 42.1 & 55.0\\
 & 896 & 16.0 & originals & 32.9 & 48.7 & +15.86 [15.48, 16.26] & 53.3 & 67.0\\
RoPE ViT-B & 224 & 1 & both & 83.7 & = & -- & -- & --\\
 & 320 & 2.0 & crops & 83.7 & 83.8 & +0.11 [0.00, 0.21] & 96.6 & 96.6\\
 & 320 & 2.0 & originals & 84.0 & 84.2 & +0.19 [0.08, 0.29] & 96.9 & 96.9\\
 & 448 & 4.0 & crops & 81.4 & 81.8 & +0.42 [0.25, 0.58] & 95.6 & 95.9\\
 & 448 & 4.0 & originals & 82.7 & 83.2 & +0.47 [0.31, 0.63] & 96.3 & 96.3\\
 & 640 & 8.2 & crops & 74.8 & 77.5 & +2.67 [2.43, 2.91] & 92.2 & 93.7\\
 & 640 & 8.2 & originals & 77.8 & 80.1 & +2.29 [2.06, 2.52] & 93.7 & 94.9\\
 & 896 & 16.0 & crops & 56.9 & 67.4 & +10.42 [10.07, 10.77] & 79.8 & 87.5\\
 & 896 & 16.0 & originals & 61.3 & 70.4 & +9.07 [8.73, 9.40] & 83.1 & 89.2\\
RoPE-Mixed ViT-B & 224 & 1 & both & 83.8 & = & -- & -- & --\\
 & 320 & 2.0 & crops & 84.0 & 84.1 & +0.09 [\textminus{}0.02, 0.20] & 96.7 & 96.7\\
 & 320 & 2.0 & originals & 84.3 & 84.5 & +0.12 [0.01, 0.23] & 97.0 & 97.0\\
 & 448 & 4.0 & crops & 82.2 & 82.9 & +0.75 [0.58, 0.91] & 96.1 & 96.4\\
 & 448 & 4.0 & originals & 83.3 & 83.9 & +0.61 [0.45, 0.77] & 96.5 & 96.7\\
 & 640 & 8.2 & crops & 75.1 & 78.6 & +3.58 [3.33, 3.83] & 92.5 & 94.6\\
 & 640 & 8.2 & originals & 78.5 & 81.3 & +2.82 [2.59, 3.05] & 94.1 & 95.5\\
 & 896 & 16.0 & crops & 49.6 & 63.2 & +13.56 [13.17, 13.95] & 73.7 & 83.3\\
 & 896 & 16.0 & originals & 56.5 & 68.1 & +11.59 [11.22, 11.98] & 79.4 & 86.9\\
\addlinespace
\multicolumn{9}{@{}l}{\emph{Registers, average pooling}}\\
ViT-M, 4 registers & 256 & 1 & both & 83.5 & = & -- & -- & --\\
 & 368 & 2.1 & crops & 82.6 & 82.7 & +0.18 [0.07, 0.30] & 96.1 & 96.2\\
 & 368 & 2.1 & originals & 83.1 & 83.4 & +0.28 [0.17, 0.39] & 96.5 & 96.6\\
 & 512 & 4.0 & crops & 79.1 & 80.1 & +0.91 [0.73, 1.09] & 94.4 & 95.0\\
 & 512 & 4.0 & originals & 81.0 & 81.9 & +0.87 [0.70, 1.03] & 95.4 & 95.8\\
 & 720 & 7.9 & crops & 72.1 & 74.2 & +2.08 [1.85, 2.32] & 90.4 & 92.2\\
 & 720 & 7.9 & originals & 75.4 & 77.6 & +2.23 [2.01, 2.45] & 92.4 & 93.9\\
 & 1024 & 16.0 & crops & 60.9 & 65.0 & +4.12 [3.82, 4.41] & 82.4 & 86.6\\
 & 1024 & 16.0 & originals & 63.5 & 67.9 & +4.45 [4.15, 4.74] & 84.5 & 88.3\\
ViT-Betwixt, 4 registers & 256 & 1 & both & 83.7 & = & -- & -- & --\\
 & 368 & 2.1 & crops & 82.6 & 82.8 & +0.21 [0.10, 0.33] & 96.0 & 96.2\\
 & 368 & 2.1 & originals & 83.4 & 83.7 & +0.34 [0.23, 0.45] & 96.5 & 96.6\\
 & 512 & 4.0 & crops & 78.9 & 79.9 & +0.94 [0.77, 1.13] & 94.0 & 94.6\\
 & 512 & 4.0 & originals & 80.8 & 81.7 & +0.91 [0.74, 1.08] & 95.0 & 95.6\\
 & 720 & 7.9 & crops & 69.4 & 72.2 & +2.83 [2.57, 3.09] & 87.8 & 89.8\\
 & 720 & 7.9 & originals & 72.8 & 75.6 & +2.81 [2.55, 3.06] & 89.8 & 91.7\\
 & 1024 & 16.0 & crops & 52.3 & 55.6 & +3.27 [2.94, 3.60] & 73.9 & 76.6\\
 & 1024 & 16.0 & originals & 56.5 & 59.9 & +3.43 [3.11, 3.76] & 77.2 & 80.0\\
\addlinespace
\bottomrule
\end{tabular}
\caption{\textbf{Every rung and route: rotary positions and registers.}}
\label{tab:sup-rungs-other}

\vspace{0.5em}
\begin{tabular}{@{}lrrrrrr@{}}
\toprule
& & \multicolumn{3}{c}{Crowd's log-sum} & & \\
\cmidrule(lr){3-5}
Model & Layers & All & First third & Last third & Entropy & Top score\\
\midrule
DeiT-S & 12 & 0.81 & 1.05 & 0.76 & 0.99 & +0.02\\
DeiT-B & 12 & 0.91 & 1.01 & 0.75 & 0.99 & +0.09\\
DeiT III-B & 12 & 0.77 & 0.97 & 0.61 & 1.00 & \textminus{}0.05\\
AugReg ViT-S & 12 & 0.90 & 1.06 & 0.73 & 0.99 & +0.12\\
AugReg2 ViT-B & 12 & 0.82 & 1.06 & 0.57 & 1.00 & +0.15\\
AugReg ViT-L & 24 & 0.90 & 0.94 & 0.83 & 0.98 & +0.08\\
\bottomrule
\end{tabular}
\caption{\textbf{The class token's slopes per unit log patch count}, as trained, on enlarged crops (medians over images, heads and
layers; the thirds are means of the layer medians).}
\label{tab:sup-spread}
\end{table*}

\begin{table*}[p]
\centering
\footnotesize
\begin{tabular}{@{}llrrrr@{}}
\toprule
& & \multicolumn{2}{c}{Measured} & \multicolumn{2}{c}{Predicted from training scores}\\
\cmidrule(lr){3-4}\cmidrule(l){5-6}
Model & Option (rows) & As trained & Rescaled & As trained & Rescaled\\
\midrule
\multicolumn{6}{@{}l}{\emph{Production models}}\\
DINOv2-S, registers & registers (class token) & \textminus{}0.448 & +0.029 & \textminus{}0.943 & \textminus{}0.119\\
DINOv2-B, registers & registers (class token) & \textminus{}0.496 & +0.055 & \textminus{}0.969 & \textminus{}0.134\\
DINOv2-L, registers & registers (class token) & \textminus{}0.350 & +0.065 & \textminus{}0.985 & \textminus{}0.130\\
ViT-M, 4 registers & registers (patches) & \textminus{}0.783 & \textminus{}0.276 & \textminus{}1.005 & \textminus{}0.518\\
ViT-Betwixt, 4 registers & registers (patches) & \textminus{}0.644 & \textminus{}0.223 & \textminus{}1.005 & \textminus{}0.499\\
\multicolumn{6}{@{}l}{\emph{Controlled ViT-S, three seeds}}\\
zero-value registers, seed 1 & registers & \textminus{}0.956 & \textminus{}0.426 & \textminus{}0.994 & \textminus{}0.363\\
zero-value registers, seed 2 & registers & \textminus{}0.965 & \textminus{}0.357 & \textminus{}0.989 & \textminus{}0.269\\
zero-value registers, seed 3 & registers & \textminus{}0.958 & \textminus{}0.280 & \textminus{}0.994 & \textminus{}0.287\\
learned sink, seed 1 & sink & -- & -- & -- & --\\
learned sink, seed 2 & sink & -- & -- & -- & --\\
learned sink, seed 3 & sink & \textminus{}0.646 & \textminus{}0.174 & \textminus{}0.961 & \textminus{}0.510\\
gate with a sink, seed 1 & gate & +0.010 & \textminus{}0.004 & -- & --\\
gate with a sink, seed 2 & gate & +0.006 & \textminus{}0.001 & -- & --\\
gate with a sink, seed 3 & gate & +0.016 & +0.012 & -- & --\\
gate with a sink, seed 1 & sink inside the gate & \textminus{}0.933 & \textminus{}0.414 & \textminus{}0.986 & \textminus{}0.577\\
gate with a sink, seed 2 & sink inside the gate & \textminus{}0.983 & \textminus{}0.265 & \textminus{}0.996 & \textminus{}0.429\\
gate with a sink, seed 3 & sink inside the gate & \textminus{}0.736 & +0.042 & \textminus{}0.995 & \textminus{}0.325\\
\bottomrule
\end{tabular}
\caption{\textbf{Log-odds slopes in $\ln N$ of the share on each option.} A dash: no heads use the option, or (for the
gate) no prediction applies.}
\label{tab:sup-shares}

\vspace{1.2em}
\begin{tabular}{@{}lrrrrr@{}}
\toprule
Model & Head threshold & Selected head/query rows & Register log-odds & Registers' log-sum-exp & Reconstructed crowd term\\
\midrule
DINOv2-S, registers & 0.05 & 61 & \textminus{}0.46 & +0.09 & 0.55\\
 & 0.10 & 47 & \textminus{}0.45 & +0.09 & 0.53\\
 & 0.20 & 30 & \textminus{}0.46 & +0.08 & 0.55\\
\multicolumn{6}{@{}r}{\footnotesize split stable across thresholds}\\
DINOv2-B, registers & 0.05 & 124 & \textminus{}0.51 & +0.09 & 0.59\\
 & 0.10 & 99 & \textminus{}0.50 & +0.09 & 0.57\\
 & 0.20 & 65 & \textminus{}0.44 & +0.09 & 0.53\\
\multicolumn{6}{@{}r}{\footnotesize split stable across thresholds}\\
DINOv2-L, registers & 0.05 & 323 & \textminus{}0.38 & +0.09 & 0.62\\
 & 0.10 & 273 & \textminus{}0.35 & +0.09 & 0.65\\
 & 0.20 & 233 & \textminus{}0.43 & +0.08 & 0.64\\
\multicolumn{6}{@{}r}{\footnotesize split stable across thresholds}\\
ViT-M, 4 registers & 0.05 & 548 & \textminus{}0.78 & +0.05 & 0.79\\
 & 0.10 & 433 & \textminus{}0.78 & +0.07 & 0.84\\
 & 0.20 & 296 & \textminus{}0.67 & +0.02 & 0.67\\
\multicolumn{6}{@{}r}{\footnotesize split stable across thresholds}\\
ViT-Betwixt, 4 registers & 0.05 & 751 & \textminus{}0.73 & +0.01 & 0.76\\
 & 0.10 & 597 & \textminus{}0.64 & +0.00 & 0.74\\
 & 0.20 & 400 & \textminus{}0.69 & \textminus{}0.05 & 0.67\\
\multicolumn{6}{@{}r}{\footnotesize split not stable across thresholds}\\
\bottomrule
\end{tabular}
\caption{\textbf{Register log-odds and their rowwise components, as trained.} Each column reports a separate median
of fitted slopes. Before aggregation, the clamped log-odds slope equals the register-group log-sum-exp slope minus the
reconstructed crowd-term slope; the medians need not satisfy that equality.}
\label{tab:sup-split}
\end{table*}

\begin{table*}[p]
\centering
\footnotesize
\begin{tabular}{@{}lrrll@{}}
\toprule
Model & px & Native images & Gain on originals [95\% CI] & Gain, originals $-$ crops [95\% CI]\\
\midrule
\multicolumn{5}{@{}l}{\emph{Absolute positions, class token}}\\
DeiT-S & 448 & 4{,}257 & +2.80 [2.04, 3.55] & \textminus{}0.07 [\textminus{}1.03, 0.87]\\
 & 640 & 1{,}019 & +7.65 [5.59, 9.81] & \textminus{}0.88 [\textminus{}3.63, 1.86]\\
 & 896 & 448 & +18.53 [14.29, 22.77] & \textminus{}3.12 [\textminus{}8.48, 2.23]\\
DeiT-B & 448 & 4{,}257 & +1.06 [0.47, 1.67] & +0.07 [\textminus{}0.70, 0.85]\\
 & 640 & 1{,}019 & +1.28 [\textminus{}0.20, 2.65] & \textminus{}0.59 [\textminus{}2.55, 1.28]\\
 & 896 & 448 & +3.57 [0.89, 6.25] & \textminus{}0.45 [\textminus{}4.46, 3.57]\\
DeiT III-B & 448 & 4{,}257 & +0.49 [\textminus{}0.02, 1.03] & \textminus{}0.05 [\textminus{}0.75, 0.70]\\
 & 640 & 1{,}019 & +0.88 [\textminus{}0.59, 2.26] & +1.18 [\textminus{}0.69, 2.94]\\
 & 896 & 448 & +0.45 [\textminus{}2.46, 3.35] & \textminus{}0.45 [\textminus{}4.24, 3.12]\\
AugReg ViT-S & 448 & 4{,}257 & +1.13 [0.45, 1.81] & +0.00 [\textminus{}0.87, 0.87]\\
 & 640 & 1{,}019 & +1.96 [0.20, 3.73] & \textminus{}1.57 [\textminus{}3.83, 0.69]\\
 & 896 & 448 & +12.05 [8.04, 16.29] & \textminus{}2.46 [\textminus{}7.37, 2.46]\\
AugReg2 ViT-B & 448 & 4{,}257 & \textminus{}0.92 [\textminus{}1.55, \textminus{}0.28] & +0.05 [\textminus{}0.82, 0.92]\\
 & 640 & 1{,}019 & \textminus{}1.86 [\textminus{}3.63, 0.00] & +1.86 [\textminus{}0.59, 4.32]\\
 & 896 & 448 & \textminus{}2.01 [\textminus{}6.03, 1.79] & +4.91 [\textminus{}0.23, 10.27]\\
AugReg ViT-L & 448 & 4{,}257 & +0.56 [0.02, 1.10] & +0.61 [\textminus{}0.09, 1.32]\\
 & 640 & 1{,}019 & +0.49 [\textminus{}0.88, 1.86] & +0.79 [\textminus{}0.98, 2.65]\\
 & 896 & 448 & +1.56 [\textminus{}1.34, 4.46] & +2.01 [\textminus{}1.79, 5.80]\\
\multicolumn{5}{@{}l}{\emph{Rotary positions}}\\
RoPE ViT-S & 448 & 4{,}257 & +0.63 [\textminus{}0.09, 1.34] & +0.21 [\textminus{}0.75, 1.17]\\
 & 640 & 1{,}019 & +3.93 [1.77, 6.08] & \textminus{}0.69 [\textminus{}3.43, 1.96]\\
 & 896 & 448 & +12.95 [8.71, 17.41] & +1.34 [\textminus{}4.02, 6.70]\\
RoPE ViT-B & 448 & 4{,}257 & +0.35 [\textminus{}0.16, 0.89] & \textminus{}0.09 [\textminus{}0.75, 0.56]\\
 & 640 & 1{,}019 & +1.57 [0.20, 2.94] & +0.20 [\textminus{}1.57, 2.06]\\
 & 896 & 448 & +8.93 [5.13, 12.72] & \textminus{}2.23 [\textminus{}6.47, 2.01]\\
RoPE-Mixed ViT-B & 448 & 4{,}257 & +0.61 [0.07, 1.17] & \textminus{}0.16 [\textminus{}0.82, 0.49]\\
 & 640 & 1{,}019 & +2.45 [0.79, 4.12] & \textminus{}1.47 [\textminus{}3.43, 0.49]\\
 & 896 & 448 & +9.38 [5.58, 13.17] & \textminus{}4.24 [\textminus{}8.93, 0.45]\\
\multicolumn{5}{@{}l}{\emph{Registers, average pooling}}\\
ViT-M, 4 registers & 512 & 2{,}001 & +0.45 [\textminus{}0.25, 1.15] & \textminus{}0.20 [\textminus{}1.10, 0.70]\\
 & 720 & 926 & +1.40 [0.00, 2.81] & \textminus{}0.65 [\textminus{}2.59, 1.30]\\
 & 1024 & 375 & +2.40 [0.00, 5.07] & +1.33 [\textminus{}2.40, 5.33]\\
ViT-Betwixt, 4 registers & 512 & 2{,}001 & +0.40 [\textminus{}0.30, 1.10] & \textminus{}0.65 [\textminus{}1.55, 0.25]\\
 & 720 & 926 & +3.02 [1.30, 4.75] & \textminus{}0.32 [\textminus{}2.48, 1.73]\\
 & 1024 & 375 & +3.20 [\textminus{}0.80, 7.20] & +1.60 [\textminus{}2.93, 6.13]\\
\midrule
Pooled, six primary & 640 & 1{,}019 & +1.73 [\textminus{}0.46, 4.37] & +0.13 [\textminus{}1.23, 1.44]\\
 & 896 & 448 & +5.69 [0.37, 12.02] & +0.07 [\textminus{}2.72, 3.01]\\
\bottomrule
\end{tabular}
\caption{\textbf{Native-detail images.} Gains are paired by image; the pooled rows resample models, then images.}
\label{tab:sup-native}
\end{table*}

\begin{table*}[p]
\centering
\footnotesize
\begin{tabular}{@{}lrrlll@{}}
\toprule
Comparison & px & Paired mean & Pointwise 95\% interval & Simultaneous interval & Holm $p$\\
\midrule
\multicolumn{6}{@{}l}{\emph{Five public models}}\\
FiTv2's form $-$ $\lam$ & 640 & \textminus{}0.14 & [\textminus{}0.51, 0.18] & [\textminus{}0.63, 0.28] & 1.000\\
YaRN's form $-$ $\lam$ & 640 & \textminus{}0.24 & [\textminus{}0.65, 0.13] & [\textminus{}0.81, 0.23] & 1.000\\
Calibrated $c^*$ $-$ $\lam$ & 640 & +0.84 & [\textminus{}0.17, 2.48] & [\textminus{}0.27, 3.23] & 1.000\\
$\lam$ $-$ $\sqrt{\lam}$ & 640 & +0.42 & [\textminus{}2.02, 2.47] & -- & --\\
FiTv2's form $-$ $\lam$ & 896 & \textminus{}0.81 & [\textminus{}2.39, 0.45] & [\textminus{}2.95, 0.74] & 1.000\\
YaRN's form $-$ $\lam$ & 896 & \textminus{}0.55 & [\textminus{}1.90, 0.47] & [\textminus{}2.39, 0.68] & 1.000\\
Calibrated $c^*$ $-$ $\lam$ & 896 & +1.53 & [0.05, 3.91] & [\textminus{}0.06, 4.92] & 0.168\\
$\lam$ $-$ $\sqrt{\lam}$ & 896 & +1.24 & [\textminus{}2.87, 4.87] & -- & --\\
\multicolumn{6}{@{}l}{\emph{Fifteen controlled checkpoints}}\\
FiTv2's form $-$ $\lam$ & 640 & +0.20 & [0.09, 0.31] & [0.05, 0.35] & 0.002\\
YaRN's form $-$ $\lam$ & 640 & +0.26 & [0.12, 0.39] & [0.07, 0.44] & 0.002\\
Calibrated $c^*$ $-$ $\lam$ & 640 & +0.20 & [\textminus{}0.06, 0.48] & [\textminus{}0.14, 0.56] & 0.529\\
$\lam$ $-$ $\sqrt{\lam}$ & 640 & +1.32 & [1.06, 1.58] & -- & --\\
FiTv2's form $-$ $\lam$ & 896 & \textminus{}0.11 & [\textminus{}0.34, 0.13] & [\textminus{}0.42, 0.22] & 1.000\\
YaRN's form $-$ $\lam$ & 896 & +0.02 & [\textminus{}0.17, 0.21] & [\textminus{}0.25, 0.28] & 1.000\\
Calibrated $c^*$ $-$ $\lam$ & 896 & +0.09 & [\textminus{}0.11, 0.29] & [\textminus{}0.17, 0.37] & 1.000\\
$\lam$ $-$ $\sqrt{\lam}$ & 896 & +1.85 & [1.50, 2.20] & -- & --\\
\bottomrule
\end{tabular}
\caption{\textbf{Each multiplier rule against $\lam$ on held-out enlarged crops}, in points of top-1. The simultaneous
intervals are conservative 99.17\% Bonferroni intervals over the six comparisons in each group; the Holm $p$-values
adjust the same family. The comparison of $\lam$ with $\sqrt{\lam}$ is outside that family.}
\label{tab:sup-rules}
\end{table*}

\begin{table*}[p]
\centering
\footnotesize
\begin{tabular}{@{}lrrrrrr@{}}
\toprule
& \multicolumn{2}{c}{Loss without rescaling} & \multicolumn{4}{c}{Gain over no rescaling}\\
\cmidrule(lr){2-3}\cmidrule(l){4-7}
Model & Enlarged & Replicas & $\lam$ & Offset & Offset and $\lam$ & $\lam$, non-local heads\\
\midrule
\multicolumn{7}{@{}l}{\emph{Controlled ViT-S, mean of three seeds}}\\
Softmax & 24.4 & 0.5 & +5.0 & \textminus{}0.2 & +4.6 & +5.0\\
Zero-value registers & 25.3 & 2.3 & +5.9 & \textminus{}0.1 & +3.6 & +5.9\\
Learned sink & 24.5 & 0.7 & +4.9 & \textminus{}0.1 & +4.3 & +4.9\\
Output gate & 24.9 & 0.0 & +5.0 & +0.1 & +5.8 & +5.0\\
Gate and sink & 24.5 & 0.0 & +5.0 & +0.3 & +5.4 & +4.8\\
\multicolumn{7}{@{}l}{\emph{Public models}}\\
DeiT-S & 48.0 & 6.1 & +21.4 & +16.4 & +23.1 & +5.8\\
DeiT-B & 22.1 & 0.8 & +7.5 & \textminus{}1.5 & +5.2 & +5.3\\
DeiT III-B & 19.0 & 2.2 & +3.1 & \textminus{}1.7 & \textminus{}2.0 & +1.9\\
AugReg ViT-S & 33.0 & 4.2 & +13.1 & +3.2 & +8.1 & +8.1\\
AugReg2 ViT-B & 38.5 & 4.5 & 0.0 & +19.6 & +19.9 & \textminus{}1.7\\
ViT-M, 4 registers & 22.4 & 2.4 & +4.1 & \textminus{}2.2 & \textminus{}2.2 & +2.2\\
ViT-Betwixt, 4 registers & 31.5 & 5.5 & +3.0 & +8.8 & +14.4 & 0.0\\
\bottomrule
\end{tabular}
\caption{\textbf{Three accounts of the rescaling's gain}, in points of top-1 at four times the training side.
\emph{Loss}: accuracy at the training size minus accuracy without rescaling, on enlarged crops and on exact token
replicas. \emph{Offset}: $\ln m$ added to the scores of the keys that enlargement does not copy. Public-model accuracies
use the evaluation half defined in \cref{sup:multipliers}. The controlled rows use the full ImageNet-100 validation set
and are means over three seeds.}
\label{tab:sup-why}
\end{table*}

\end{document}